\documentclass[journal]{IEEEtran}
\ifCLASSINFOpdf

\else

\fi

\usepackage[sanserif,full]{complexity}
\usepackage[ruled,vlined,linesnumbered]{algorithm2e}
\makeatletter
\def\BState{\State\hskip-\ALG@thistlm}
\makeatother
\usepackage{amsmath}
\usepackage{tabularx}
\usepackage{notoccite}
\usepackage{hyperref}
\usepackage{booktabs}
\usepackage{graphicx}
\usepackage{dblfloatfix}
\ifCLASSOPTIONcompsoc
 \usepackage[caption=false,font=normalsize,labelfont=sf,textfont=sf]{subfig}
\else
 \usepackage[caption=false,font=footnotesize]{subfig}
\fi

\usepackage{tikz,xcolor}

\definecolor{lime}{HTML}{A6CE39}
\DeclareRobustCommand{\orcidicon}{%
	\begin{tikzpicture}
	\draw[lime, fill=lime] (0,0) 
	circle [radius=0.16] 
	node[white] {{\fontfamily{qag}\selectfont \tiny ID}};
	\draw[white, fill=white] (-0.0625,0.095) 
	circle [radius=0.007];
	\end{tikzpicture}
	\hspace{-2mm}
}

\foreach \x in {A, ..., Z}{%
	\expandafter\xdef\csname orcid\x\endcsname{\noexpand\href{https://orcid.org/\csname orcidauthor\x\endcsname}{\noexpand\orcidicon}}
}

\begin{document}

\title{Modeling Influencer Marketing Campaigns \\ in Social Networks}

\author{Ronak~Doshi*\orcidC{}, 
         Ajay~Ramesh*\orcidB{},~\IEEEmembership{Student Member,~IEEE,} 
         and~Shrisha~Rao\orcidA{},~\IEEEmembership{Senior Member,~IEEE} 
\thanks{R. Doshi, A. Ramesh and S. Rao is with the International Institute of Information Technology Bangalore, Bengaluru 560100, Karnataka, India (e-mail: \href{mailto:ronakdoshi36@gmail.com}{ronakdoshi36@gmail.com}, \href{mailto:ranganathajay@gmail.com}{ranganathajay@gmail.com}, \href{mailto:shrao@ieee.org}{shrao@ieee.org})}
\thanks{*These authors contributed equally to this work.}
}

\markboth{IEEE Transactions On Computational Social Systems }%
{Shell \MakeLowercase{\textit{et al.}}: Bare Demo of IEEEtran.cls for IEEE Journals}

\maketitle

\begin{abstract}
Social media are extensively used in today's world, and facilitate quick and easy sharing of information, which makes them a good way to advertise products. Influencers of a social media network, owing to their massive popularity, provide a huge potential customer base. However, it is not straightforward to decide which influencers should be selected for an advertizing campaign that can generate high returns with low investment. In this work, we present an agent-based model (ABM) that can simulate the dynamics of influencer advertizing campaigns in a variety of scenarios and can help to discover the best influencer marketing strategy. Our system is a probabilistic graph-based model that provides the additional advantage to incorporate real-world factors such as customers' interest in a product, customer behavior, the willingness to pay, a brand's investment cap, influencers' engagement with influence diffusion, and the nature of the product being advertized viz. luxury and non-luxury. 
Using customer acquisition cost and conversion ratio as a unit economic, we evaluate the performance of different kinds of influencers under a variety of circumstances that are simulated by varying the nature of the product and the customers' interest. Our results exemplify the circumstance-dependent nature of influencer marketing and provide insight into which kinds of influencers would be a better strategy under respective circumstances. For instance, we show that as the nature of the product varies from luxury to non-luxury, the performance of celebrities declines whereas the performance of nano-influencers improves. In terms of the customers' interest, we find that the performance of nano-influencers declines with the decrease in customers' interest whereas the performance of celebrities improves.   
\end{abstract}
\begin{IEEEkeywords}
Agent Based Modeling, Influencer Marketing, Social Networks, Viral Marketing
\end{IEEEkeywords}

\IEEEpeerreviewmaketitle

\section{Introduction}
\IEEEPARstart{S}{ocial} media usage has been significantly rising over the years. Social networking platforms are now a primary stage for discussion, discourse, opinion and information sharing, recreational activities, etc. As a result, they are extensively used in a variety of domains such as business, education, medicine, finance, and politics. Online social networking platforms and their vast and diverse user population offer significant opportunities for marketers to engage with customers. Moreover, certain individuals in social networks are hugely popular and influential. These \emph{influencers} can alter the opinions and ideas of their followers \cite{cercel-opinion-propagation}, and thus possess marketing potential which is much sought after by brands to endorse their products to the audience in the network. The social influence exerted by these influencers is known to play a significant role in the success of advertizing and marketing \cite{cicvaric}. This sort of marketing is called \emph{influencer marketing}.

Influencer marketing has become one of the fastest and most effective marketing techniques since the inception of online social networks. A recent survey by Rakuten advertizing \cite{rakuten-advertising} showed that around $80\%$ of consumers made purchases recommended by influencers. A social network influencer can promote a product/brand to a huge audience with a click of a button. Studies have shown that partnering with influencers for advertizing results in an increase in a brand's revenue within a short period of time \cite{casestudy}. However, there are several complexities involved in implementing influencer marketing. It is not straightforward to select the best set of influencers who can generate the highest returns. An influencer marketing survey by Mediakix \cite{IM-difficult-survey} showed that $61\%$ of the marketers surveyed agreed that it is difficult to find the right influencers for campaigns. Additionally, in recent times, influencers are found in different kinds ranging from celebrities to nano-influencers (in terms of the number of followers they have). Experimental studies have shown that even nano-influencers play significant roles in advertizements \cite{casestudy}\cite{rise-of-micro} and the conventional expectation that celebrities are always the best choice for advertizing is incorrect. This  is  primarily  because nano-influencers  actively  engage  with  their  followers which occurs through comments, replies, etc. This engagement with potential customers has a substantial effect on marketing results. Thus, unlike in the past when a marketer would just employ a big influencer to advertize a product, the present online social network advertizing strategies depend on a variety of network and brand-specific factors. Network-specific factors include the customers' interest in the product, the kinds of influencers and their engagement in the network, and the structure of the network itself. Brand-specific factors include the hiring investment, the nature of the product being advertized, and the marketing goals.

Several works exist regarding the diffusion or propagation of influence in networks. A seminal work in the field was by Kempe \textit{et al.} \cite{Kempe} who proposed two kinds of models\textemdash \textit{linear threshold} (LT) and \textit{independent cascade} (IC). These models have also been used for various applications such as voting, opinion spread, etc \cite{holley-voter}\cite{fushimi-voter}.
However, there exist only a few works that model social network marketing campaigns in particular. Among these are Wang \textit{et al.} \cite{wang-2014}\cite{wang-2015a} \cite{wang-2018}, and Domingos and Richardson \cite{domingos} who propose models in the context of viral marketing. Agent-based modelling approaches include \textemdash word-of-mouth marketing \cite{brudermann}\cite{wordofmouth}, and modelling of opinion dynamics \cite{abms-influence}. But, these existing models do not incorporate factors specifically pertinent to influencer marketing. For instance, they do not account for the new paradigm in influencer marketing involving the different kinds of influencers. Although these models effectively incorporate the diffusion of influence and opinions in the network, they do not incorporate customer behavior which plays a major role in an individual's interactions and the decision-making process. The state of art is described in further detail in Section \ref{section_bg}.

In this work, we modify and build upon the basic models mentioned above and present the design of an agent-based model (ABM) to simulate influencer advertizing campaigns in social networks. The social graph with agents at vertices forms our base framework. We incorporate real-world characteristics in the form of novel agent and model parameters that are essential in the influencer marketing context but are absent in the current literature. These include the different kinds of influencers with their respective engagements \cite{engagement-socialmedia-advertising}, the customers' interest in the product, and the customers' willingness to pay. The interactions of individuals in the social network are modeled as two types of agent behavior\textemdash influencer-follower and inter-follower interactions. 

We use our model to study the performance of different kinds of influencers in a social network under varying circumstances, which has also not been attempted before. Given an initial set of influencers, the model simulates a campaign in a breadth-first manner, which in reality, corresponds to an influencer using a social network channel (a post in the form of an image, video, etc.) to advertize a product to their followers. These followers in turn propagate it to their sub-networks/followers (by sharing, retweeting, etc.). To simulate a specific advertizing plan/campaign, a social network of choice in the form of a social graph, the corresponding customers' interest and willingness to pay, the influencer hiring and engagement rates of the network, etc., can be provided as inputs to the model.

The rest of the paper is structured as follows: In Section \ref{section_bg}, we review relevant literature, the state of the art, and describe influencer marketing and its challenges. Then we present the design of our ABM in Section \ref{section_design} followed by experimental results in Section \ref{section_exp}. We conclude our work in Section \ref{section_conc}, and present additional results in Appendix \ref{apdx:appendix}. 

\section{Background}
\label{section_bg}
Several attempts have been made to model the processes by which ideas, opinions, and influence propagate through social networks. Granovetter \cite{granovetter} introduced the notion of \textit{thresholds}: the number or proportion of others who must make one decision before a given actor does. Threshold models can be applied in many scenarios such as the diffusion of innovation, the spread of diseases, and in situations where actors are required to take binary decisions \cite{granovetter}. Kempe \textit{et al.} \cite{Kempe} generalize this concept and propose two influence diffusion models for the spread of ideas in social networks\textemdash \textit{linear threshold} (LT) and \textit{independent cascade} (IC). The diffusion is carried out in discrete steps wherein each agent is activated when the combined influence of its neighbors is greater than a particular threshold possessed by the agent. In addition, Kempe \textit{et al.} \cite{Kempe} consider the problem of influence maximization: finding the set of most influential nodes in the network. They show that this problem is $\NP$-hard and propose algorithms to obtain good solutions. Various improvements have also been made to the LT and IC models. Barbieri \textit{et al.} \cite{barbieri} propose \textit{topic aware} extensions in which  authoritativeness, influence, and relevance are explicitly incorporated into the LT and IC models. 
Guo \textit{et al.} \cite{guo} proposed a \textit{greedy hill-climbing algorithm} to solve influence maximization for complementary products.

Various models for the diffusion of opinion, in particular, have been proposed. These include the voter model \cite{holley-voter}\cite{fushimi-voter}\cite{Suchecki}, the Deffaunt model \cite{deffaunt}, and the Sznajz model \cite{rodrigues-snaz}. The concept of an \emph{opinion leader} was introduced by Lazarsfeld \cite{lazarsfeld}. Opinion leaders play a central role in influencing opinion and behavior. But, identifying them is a challenge \cite{sharara-opinion-leaders}. Other literature tries to identify influential nodes in a graph or ``opinion leaders'' in a social network, and uses techniques based on graph centrality \cite{berahmand}\cite{BERAHMAND201841} for this purpose.  However, our work is not concerned with identifying specific individuals as influencers, but aims to analyze how using influencers big and small affect advertising campaigns. 

Wang \textit{et al.} \cite{wang-2014} introduces a new influence cascade model and a novel algorithm for community detection and influence ranking in social networks. In their subsequent work \cite{wang-2015a}, they use their influence model to uncover influence centrality and community structure in social networks.  Our model is inspired mainly by this work.

In the context of viral marketing, Domingos and Richardson \cite{domingos} propose a model for a customer's network value, which includes the expected sales from the customer and others who are influenced to become customers. They view customers in the market as nodes of the social network and model their influence on each other as a Markov random field. In their subsequent work \cite{richardson}, they pose the problem of finding the best marketing plan for a given social network and propose a simple linear model to do so. Wang \textit{et al.} \cite{wang-2018} build upon their previous work \cite{wang-2014} to model and maximize influence diffusion in social networks for viral marketing.     

\section{Model Design}
\label{section_design}
Our system is a probabilistic graph-based model inspired by the Independent Cascade (IC) model \cite{Kempe}. We model every individual in the social network as an agent that exerts influence on its followers and is in turn influenced by the agents that it follows. The model is integrated into a weighted directed social graph $G$, wherein each agent occupies a vertex in the graph. A directed edge going from $x$ to $y$  $\left(x,y\right)$ represents a connection between two agents $x$ and $y$ where agent $y$ is a follower of agent $x$ in the social network. The social influence of agent $x$ on agent $y$ is commonly defined as the power that $x$ has to effect a change in the opinion of $y$ \cite{cercel-opinion-propagation}\cite{peng-survey}. This influence of $x$ over $y$ is represented by the edge-weight $w_{x,y}$ and interpreted as the probability of agent $x$ persuading agent $y$ to purchase a product. The graph is thus defined as follows 
\begin{equation}
    G = (V, E, W)
\end{equation}
where: 
\begin{itemize}
    \item $V$, a set of vertices/agents.

    \item ${\displaystyle E\subseteq \left\{(x,y)\mid (x,y)\in V^{2}\;{\textrm {and}}\;x\neq y\right\}}$
    
    \item $W\subseteq \left\{w_{x,y}\mid x,y\in V\;{\textrm {and}}\;x\neq y\right\}$ where, $0\leq w_{x,y}\leq 1$
\end{itemize} 
In this framework, every agent in the network is considered an influencer of some degree. We model crucial real-world behavior in the form of two kinds of agent interactions\textemdash influencer-follower and inter-follower interactions. We model the nature of the product being advertized using the notion of willingness to pay and model the customers' interest by an agent's interest parameter and its associated behavior.  

The model parameters, agent attributes, agent behavior and marketing campaign propagation are described in detail in subsections \ref{subsection_parameters}, \ref{agent}, \ref{subsection_behavior}, and \ref{subsection_propagation} respectively.

\subsection{Model Parameters}
\label{subsection_parameters}
The model parameters are essentially the inputs to the model and are provided according to the marketing scenario which is to be simulated. 

\textbf{interest ($\lambda_{x}$):} A potential customer could gain interest in a product in many ways such as having heard of it from a friend, personal interest in the topic that the product is about (e.g. sports in case of sports equipment), necessity, etc. It has been shown that this kind of consumer psychology is one of the factors that affect a potential customer's interest in a product and subsequent possible purchase \cite{donny-2018}. We thus incorporate this notion of customers' interest with the parameter $\lambda$. It can be interpreted as the probability of an agent being interested in the product, with $1$ showing maximum interest and $0$ indicating no interest. The overall interest of the population in a product is modeled as a Gaussian distribution with a certain mean ($\mu$) and variance ($\sigma^{2}$). 
    A marketer might not know the interest of every agent, but have a general idea of the overall customers' interest of the population and can set the initial mean ($\mu$) and variance ($\sigma^{2}$) of the customers' interest before the simulation. Then, the initial interest of each agent is assigned by sampling from this distribution. A high value of $\mu$ is used to simulate a scenario where the customers' interest in the product is high and a low value of $\mu$ is used to simulate a scenario where the customers' interest in the product is low. The interest of an agent is dynamic and changes according to its interactions in the network. These interactions are later defined in Section \ref{agent}. 
    
\textbf{fraction of network willing to pay ($\Omega$): } The willingness to pay is the maximum price above which a consumer will definitely not buy one unit of a product \cite{varian}. The willingness to pay has been shown to depend on various factors such as altruism, price consciousness, reference price, income, and perceived fairness \cite{rajat-willingness-factors}. A marketer may not know the exact willingness to pay of every agent in the network but can estimate the willingness to pay for a product and approximate the fraction of the network that is willing to pay the price for a product \cite{willingnesscal}. We incorporate this fraction by the parameter $\Omega$. This parameter is intended to capture the nature of the product being advertized, in terms of its price\textemdash non-luxury or luxury. We make a correlation between the nature of the product and the willingness fraction based on a reasonable assumption that the number of people willing to pay for a high-priced product will be less than the number of people willing to pay for a low-priced product. Thus, low values of $\Omega$ are used to simulate luxury products (e.g. a luxury car) and high values are used to simulate non-luxury products.     

\textbf{influence attenuation coefficient ($\alpha$):} When social influence diffuses during a marketing campaign, the rate of spread of influence is not constant and has been found to decrease with depth \cite{networks-crowds-book}. As a result, a marketing campaign cannot run endlessly and will decay after a certain duration. We capture this phenomenon by quantifying the influence attenuation along a path by a \textit{depth associated attenuation coefficient} ($\alpha$) which is used in the \textit{reachability-based} influence diffusion model proposed by Wang \textit{et al} \cite{wang-2014}\cite{wang-2015a}.
    \begin{equation}
        \alpha_{d} = d^{-2}
    \end{equation}
where $d$ is the depth (number of hops) from an influencer to the node of interest. It can be interpreted as the probability of an influencer’s influence reaching a node $d$ hops away, as analogous to the probability of a center node linking to a node at a fixed distance $d$ of its concentric scales of resolution, which is proportional to $d^{-2}$ as depicted by Easley and Kleinberg \cite{networks-crowds-book} \cite{wang-2018}. The effective influence received by a node $x$ which is at depth $d$ from the source of the advertizement campaign, by one of its neighbors $y$ is calculated as 
    \begin{equation}
        \bar{w}_{x,y} = w_{x,y} \cdot \alpha_{d}
    \end{equation}
where $\bar{w}_{x,y}$ is the effective influence of agent $y$ on $x$ and $w_{x,y}$ is the original influence of $y$ on $x$.

\textbf{hiring investment ($\rho$) : } The maximum investment that a marketer is ready to make to hire a set of influencers.

\subsection{Agent Attributes}
\label{agent}
The agent's attributes have been designed to realistically represent an individual in a social network. They are described below.

\textbf{engagement rate ($\epsilon_{x}$): } The notion of \textit{engagement} has been extensively studied in a variety of networks \cite{peng-survey}\cite{engagement-effects}. Engagement in social media networks occurs via comments, likes, retweets, etc., and influencers' engagement with potential customers has a significant impact on social media advertizing \cite{engagement-socialmedia-advertising}. We incorporate an episode of engagement, also called a \emph{stroke} \cite{berne}, which has been quantified and used before in social networks \cite{koley}. In the case of influencer marketing, it is called the engagement rate $\epsilon_{x}$ and it measures the level of interaction by followers from content created by an influencer agent $x$. The engagement rate is hence an influencer-follower interaction. It is calculated as the ratio of total interactions in terms of the number of likes, comments, etc. to the total number of followers. It is also interpreted as the percent probability of the influencer engaging with a follower. Studies show a correlation between engagement rate and the number of followers. Engagement rates of influencers generally decrease with the increase in the number of followers. In other words, celebrities have much lower engagement rates than nano and micro-influencers. The impact of the engagement rate $\epsilon$ on the outcome of an influencer's advertizing campaign will be explicated in the experimental results.

\textbf{activeness ($a_{x}$): }No member of a social network is active all the time. Inactive members do not propagate an advertizement campaign to their sub-networks, nor do they interact with other agents, i.e., they are \textit{passive}. We borrow this notion of passivity from Romero \textit{et al.} \cite{romero-passivity} and define a simple activeness attribute that represents the probability that an agent is active at a particular step in the simulation. Active networks can be simulated using high values of $a$ and less active networks using lower values.

\textbf{hiring cost ($h_{x}$): }The cost of hiring an influencer for an advertizing campaign in the network depends on the influencer's follower count. We use an estimate of approximately \$10 per 1000 followers \cite{hiringcost}. The hiring cost of an agent $x$ is shown in Equation \eqref{eq:hiring_cost}.
     \begin{equation}
     \label{eq:hiring_cost}
         h_{x} = 0.01 \times f_{x}
     \end{equation}
     where $f_{x}$ is the number of followers of the agent $x$.

The interactions and behavior of an influencer and a consumer during an advertizing campaign depend on the above mentioned attributes. These are described next.

\subsection{Agent Behavior}
\label{subsection_behavior}
An agent's behavior during an advertizing campaign comprises of interactions and a decision-making process. Interactions in social networking campaigns occur in multiple ways such as direct messaging, sharing and forwarding content, commenting on content, etc. An individual's decision to purchase a product is influenced by their interactions in the network. As mentioned before, agent interactions occur in two kinds\textemdash influencer-follower and inter-follower. The decision-making process of an agent and the outcome of its interactions with other agents are mathematically described below:

\subsubsection{Purchase Decision} 
When a campaign reaches an agent $x$, the agent has the capability to decide whether or not to purchase the product. Purchasing a product is considered as a means of participating in the advertizing campaign and deciding not to purchase the product means that the agent is unlikely to participate further in the campaign, since in reality, individuals do not generally share, retweet, or forward product-related content if they are uninterested \cite{donny-2018}. The decision to purchase depends on the interest that an agent has in the product/campaign and the influence exerted by the influencer who propagated the campaign to the agent. This decision making process is expressed by Equation \eqref{eq:decision_eq}.  
\begin{equation}
    P(\zeta_{x} = \textit{Buy Product}) = \lambda_{x}\cdot \bar{w}_{y,x} = \lambda_{x}\cdot w_{y,x} \cdot \alpha_{d}
    \label{eq:decision_eq}
\end{equation}
Where the event of an agent $x$ purchasing the product is denoted by $\zeta_{x}$. The probability of purchasing a product is computed as the product of the probability of an agent $x$ being interested in the product ($\lambda_{x}$) and influencer agent $y$'s effective influence ($\bar{w}_{y,x}$). The effective influence is a result of the existing influence ($w_{x,y}$) being attenuated by the \textit{influence attenuation coefficient} $\alpha_{d}$ at a depth of $d$ from the source of the campaign. This is similar to how Wang \textit{et al.} quantify the influence along a diffusion path \cite{wang-2015a}.     

\subsubsection{Interactions} 
An agent interacts either with an influencer who is advertizing a product or with its fellow followers in the network. The influencer-follower interaction involves the engagement of an influencer with its follower which results in an influence update. The inter-follower interaction involves agents interacting with each other and results in an update of their respective interests in the product. These interactions are described below:     
\begin{itemize}
    \item \label{item:engagment_intereaction} The engagement of an influencer with its followers has already been quantified in \ref{agent} in terms of the \emph{engagement rate}. An influencer who is advertizing a product engages with its followers according to its engagement rate. As a result, the influence received by a follower will increase. This influence update is shown in Equation \eqref{eq:engage_eq}. The probability of this engagement occurring is defined as shown in Equation \eqref{eq:eng_probability}
    \begin{equation}
        w_{x,y} = w_{x,y} + c
        \label{eq:engage_eq}
    \end{equation}
    with 
    \begin{equation}
        P(\text{influence update}) = \frac{\epsilon_{x}}{100}
        \label{eq:eng_probability}
    \end{equation}
    Where the influence (edge weight) of an advertizing influencer $x$ on a follower $y$ increases by a constant $c$ as a result of engagement. 
    \item \label{item:interest_intereaction} If an agent $x$ decides to purchase the product, it positively interacts with its followers and conveys its decision or opinion of the product. In reality, this happens via retweets, shares, comments, messaging word of mouth, etc. As a result, there is an increase in interest among followers. This is shown in the Equation \eqref{eq:interestinc}.
    \begin{equation}
        \lambda_{y} = \lambda_{y} + (w_{x,y}\cdot \lambda_{y})
        \label{eq:interestinc}
    \end{equation}
    If agent $x$ buys the product and agent $y$ is its follower. The interest of the follower agent $y$ increases according to its own initial interest $\lambda$ and the influence incoming from agent $x$ $w_{x,y}$.
    
    \item If an agent  $x$ does not purchase the product, it interacts with its followers with a low probability $\gamma$, resulting in a decrease of interest in the followers. A low probability is chosen because in general people do not interact through retweets, shares, comments, messaging or word of mouth when they do not purchase a product or ignore an advertizement campaign. The interest update is performed as shown in the Equation \eqref{eq:interestdec}.
    \begin{equation}
        \lambda_{y} = \lambda_{y} -  (w_{x,y}\cdot \lambda_{y})
        \label{eq:interestdec}
    \end{equation}
    with
    \begin{equation}
        P(\text{Interest Decrease}) = \gamma
        \label{eq:interestdec2}
    \end{equation}
\end{itemize}

We have described the attributes, parameters and behaviour of the agents in the our model. This is summarized in Table \ref{tab:parameters}. We next describe how we simulate an advertizing campaign.
\begin{table}
  \caption{Agent based model's parameters and attributes}
  \label{tab:parameters}
  \begin{tabular}{cl}
    \toprule
    Network & Description\\
    \midrule
    $G$ & Social network graph \\
    $V$ & Set of vertices of the graph \\
    $E$ & Set of edges of the graph \\
    $\nu_{i}$ & Normalized out-degree of vertex $i$ \\
    \midrule
    Model Parameters    \\
    \midrule
    $\lambda_{x}$ & Probability agent $x$ interested in product\\  
    $\Omega$ & Fraction of network willing to pay\\
    $\rho$ & Brand's influencer hiring investment \\
    $\epsilon_{x}$ & Engagement rate of the agent $x$ \\
    $a_{x}$ &  Probability of an agent $x$ being active \\
    $\gamma$ & Probability agent propagates negative influence \\
    $h_{x}$ & Hiring cost of the agent $x$ \\
    \midrule
    Model Attributes & \\
    \midrule
    $\alpha$ & Diffusion attenuation factor \\
    $w_{x,y}$ & Influence of agent $x$ on agent $y$ \\
    $f_{x}$ & Number of followers (outdegree) of the agent $x$\\
    $\zeta_{x}$ & Purchase decision of agent $x$ \\
    $\eta$ & Total hiring cost of a set of influencers\\
    \midrule
    Evaluation Metrics & \\  
    \midrule
    $\xi$ & Customer Acquisition Cost \\
    $\Theta$ & Conversion Ratio \\
    \bottomrule
  \end{tabular}
\end{table}

\subsection{The Marketing Campaign Propagation}
\label{subsection_propagation}
The simulation of a marketing campaign using our model involves the propagation of the campaign through the network from a set of source influencers. Depending on the network and nature of the campaign a marketer wishes to simulate, the input parameters of the model, viz. the network graph, the hiring rates of influencers in this network, the mean interest of the network, and the engagement rates of influencers in the network can be provided. The initialization and propagation of a marketing campaign described below: 

\subsubsection{Initialization}
 We first initialize the social network graph wherein each vertex is an agent. We then initialize the agent's parameters: the interest $\lambda_{x}$ of an agent is initialized by sampling from the truncated normal distribution according to the input mean $\mu$ and variance $\sigma^{2}$ parameters, i.e., $\lambda_{x}$ $\sim N(\mu,\sigma^{2})$ in order to obtain values in the interval [0, 1]. An agent is assigned an influencer type according to its follower count with respect to the maximum outdegree in the network (Equation \ref{eq:follower_range}). Then each agent is assigned the engagement rate corresponding to its influencer type. The types of inflencers and their corresponding engagement rate is shown in Table \ref{tab:engagementtable}.
 \begin{equation}
     \label{eq:follower_range}
     \nu_{i} = \frac{\text{Outdegree (follower count) of agent \textit{i}}}{\text{Maximum outdegree in the entire network}} \times 100
 \end{equation}
 
 Finally, the agent-to-agent influence value, i.e., the weight of the directed edge is assigned randomly by sampling from a uniform distribution, $w_{x,y} \sim U(0,1)$. We assign edge weights randomly in this work to present generic results. However, if one possesses the network graph along with edge weights that represent influence, they can be assigned accordingly. 

\begin{table}[h!]
  \centering
  \caption{Influencer types and their engagement rate. Celebrities are the biggest influencer type and nano-influencers are the smallest.}
  \label{tab:engagementtable}
  \begin{tabular}{|c |c| c |c|}
    \hline
    \textbf{Level} & \textbf{Influencer} &  \textbf{Follower} & \textbf{Engagement} \\
    & \textbf{Type} & \textbf{Range} & \textbf{Rate}\\
    \hline
    1 & Celebrities & $90\% \leq \nu_{i} \leq  100\%$ & $1\%$  \\
    \hline
    2 & Mega-influencers & $50\% \leq \nu_{i} <  90\%$ & $5\%$ \\
    \hline
    3 & Macro-influencers & $25\% \leq \nu_{i} <  50\%$ & $12\%$ \\ 
    \hline
    4 & Mid-tier influencers & $12\% \leq \nu_{i} <  25\%$ & $18\%$ \\
    \hline 
    5 & Micro-influencers &  $6\% \leq \nu_{i} <  12\%$ & $25\%$ \\
    \hline 
    6 & Nano-influencers & $0\% \leq \nu_{i} <  6\%$ & $30\%$ \\
    \hline
\end{tabular}
\end{table}

\subsubsection{Breadth-First Propagation}
Once the model initialization is complete, we initiate the influencer advertizing campaign from a selected set of influencers. The number of influencers is constrained by the brand's investment cap ($\rho$). The selected set of influencers advertize the product to their followers. Based on their types, the influencers engage with particular followers, which results in an update of the influence exerted over them (Equation \eqref{eq:engage_eq}). In reality, this occurs through methods facilitated by the specific social network (e.g., an Instagram or Facebook advertizement post followed by comments, likes, etc.). All the influencers' \textit{active} follower agents then encounter the campaign and formulate their decision (Equation \eqref{eq:decision_eq}) to either engage with the campaign (by buying the product) or not. 

According to its decision, a follower agent interacts with its neighbors resulting in an interest update (Equation \eqref{eq:interestinc} or \eqref{eq:interestdec}) among the neighbors. Thereafter all the follower agents who purchased the product propagate the campaign to their followers (recall that only \textit{active} agents perform interactions). The same process repeats for the agents who now encounter the campaign. Thus, the campaign is simulated in waves, much like how it progresses in a social network. The campaign wave diminishes when at a certain stage there are no potential buyers, resulting in nobody to further propagate the campaign. This decay which naturally occurs in social media campaigns is captured by the \textit{influence attenuation coefficient} $\alpha$. This model of propagation also incorporates the possibility of an individual encountering a campaign through multiple diffusion paths at different time instances. In other words, the agent might not purchase the product for the first few times but might choose to do so later. This is often the case because what matters is not just an advertizement campaign reaching an individual, but also through whose influence it has reached them. For simplicity, we assume that an agent can purchase a product at most once \cite{Kempe}. 

\begin{algorithm}[!htb]
\caption{Marketing Campaign Propagation}
\label{alg:propogation}
$G(V,E,W) := \text{Social Network } \textit{graph}$ \\
$F := \text{Agent Purchase Decision}$\\ 
$P := \text{Agent influencer-follower Interaction}$ \\
$H := \text{Agent inter-follower Interaction}$\\
$q := \text{Breadth First Queue }$\\
$q.enqueue(\text{Set of choosen Influencers})$\\
\While{$q$ not empty}{
    $influencer \gets q.dequeue()$
    \For{\textit{follower (f)} \text{of} \textit{influencer (i)}}{
    $w_{i,f} \gets P(\epsilon_{i}, \ w_{i,f})$ \tcp{influence update according to the influencer's engagement rate} 
        \If{\textit{follower} \text{is active} \textbf{\textit{and}} \text{is willing to pay} \textbf{\textit{and}} \text{has not bought product}}{
            $decision \gets F(\lambda_{f}, \ \alpha_{d}, \ w_{i,f})$\\
            $H(follower, \ decision)$ \tcp{follower interacts with its neighbours according to the purchase decision}
            \If {\textit{decision} == \text{product bought}}{
                $q.enqueue(follower)$
            }
        }
    }
}
\end{algorithm}

The algorithm which propagates the advertizing campaign in the manner discussed above is shown in Algorithm \ref{alg:propogation}. In line 1, we initialize the social network graph with agents as vertices. We denote the agent purchase behavior, agent influencer-follower and inter-follower interactions in lines 2, 3 and 4 with function $F$, $P$ and $H$ respectively. In line 5-6, we initialize the breadth-first queue with the chosen set of influencer agents. In line 11, according to whether influencer engages with follower or not, there is an influence update (influencer-follower interaction). Every follower agent of a chosen influencer formulates the decision whether to buy the product or not according to line 13. Then, in line 14, the follower according to the decision interacts with its neighbors and updates their interest (inter-follower interaction). In Line 15 and 16, we add the agent to the queue if it has decided to purchase the product and the campaign propagates through this agent in the same manner described above.

The data and source code are available on open source GitHub repository \footnote{\href{https://github.com/ronak66/ABM-Influencer-Follower-Advertising-Framework}{https://github.com/ronak66/ABM-Influencer-Follower-Advertising-Framework}}.

\section{Experiments and Results}
\label{section_exp}
\begin{figure}[!t]
\centering
\subfloat[Twitter social network\label{fig:twitter_dist}]{\includegraphics[width=3in]{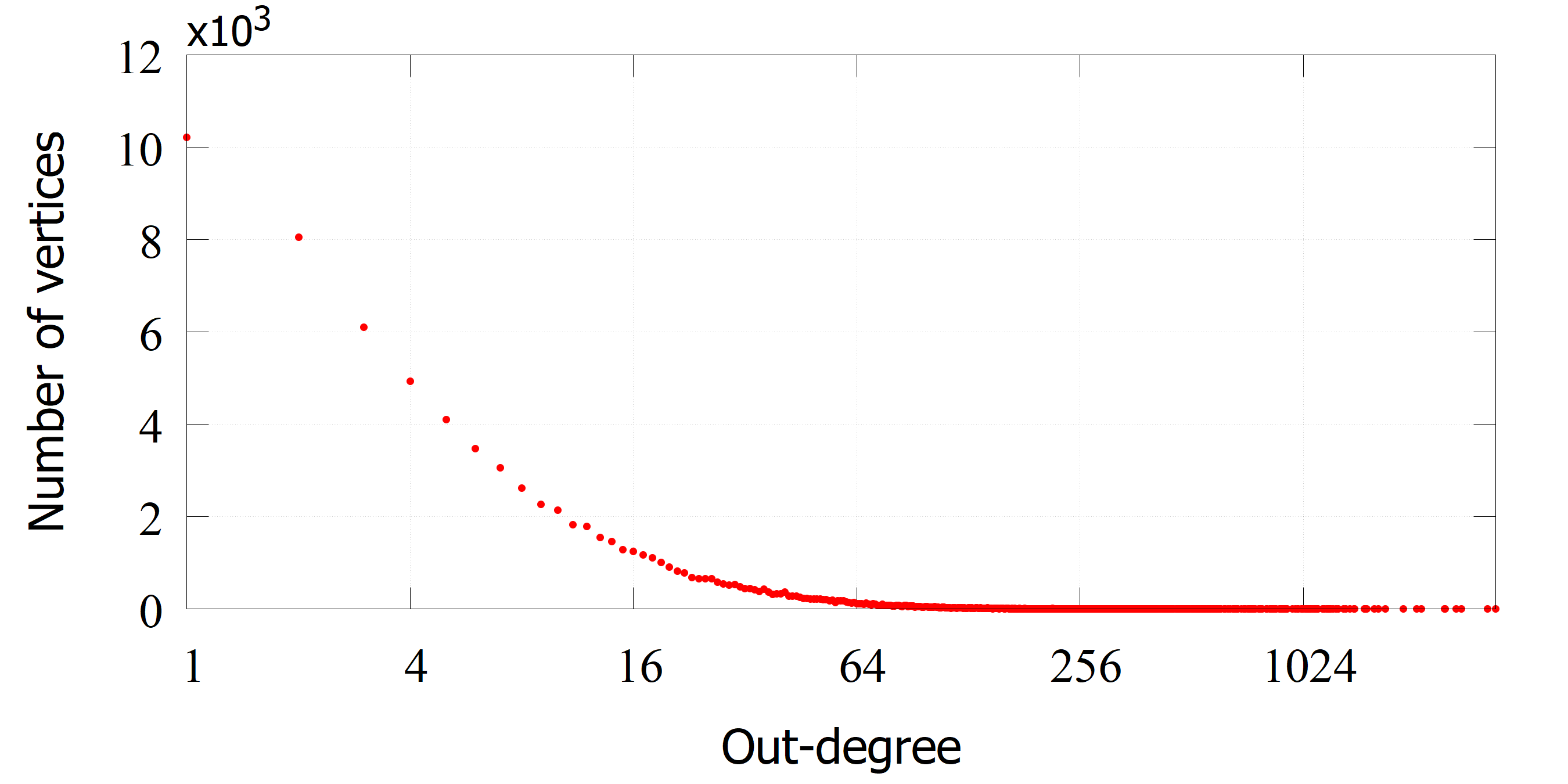}}\qquad
\hfil
\subfloat[Synthetic small world network\label{fig:syn_dist}]{\includegraphics[width=3in]{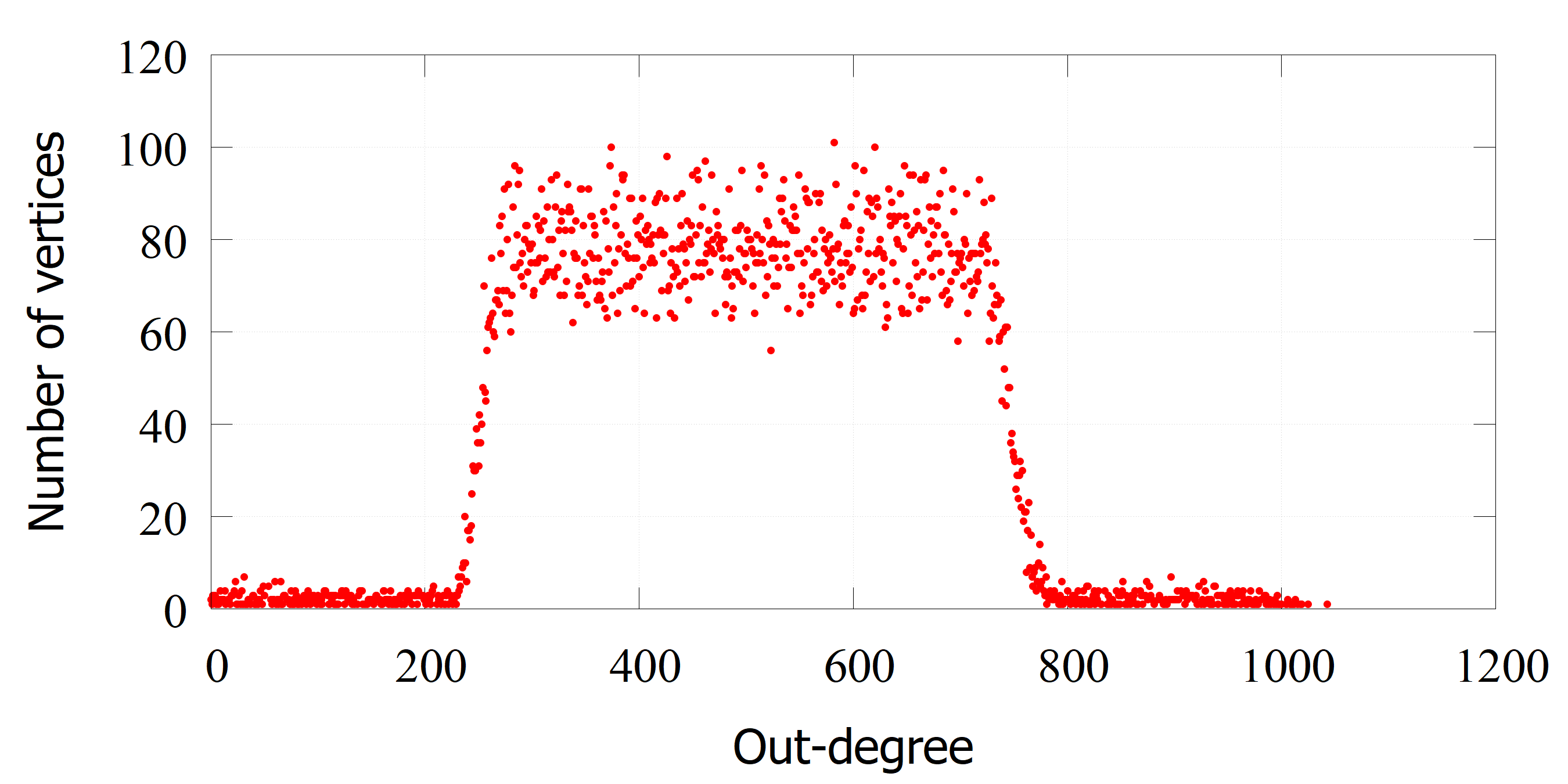}}\qquad
\caption{Out-degree distribution of the datasets}
\label{fig:outdegree_dist}
\end{figure}

Experiments were conducted to validate the working of the model and study different marketing situations. Two datasets were used for experimentation: A real-world Twitter social graph formed from social circles in Twitter \cite{snap-twitter} and a synthetic small-world graph generated using the NetworkX software \cite{networkx}. The distribution of links in real-world social networks obeys a power law \cite{networks-crowds-book}. This can be observed in Figure \ref{fig:twitter_dist}, which shows the outdegree distribution of the Twitter social graph. The synthetic small-world graph, which was generated using the Watts Strogatz model \cite{watts-model} does not obey this law, as can be seen from its outdegree distribution shown in Figure \ref{fig:syn_dist}. We experiment on both kinds of graphs and compare their results for different situations. Relevant details for the two graphs are shown in Table \ref{tab:dataset_stats}. Additionally, experimental results on a larger real-world dataset, a GooglePlus social graph \cite{snap-twitter} are presented in Appendix \ref{apdx:appendix} to validate that our model generalizes well and can be used on any real-world social network. 

\begin{table}[h!]
  \centering
    \caption{Details of the Datasets}
    \label{tab:dataset_stats}
    \begin{tabular}{|c|c|c|c|}
    \hline
     & Twitter & Synthetic & Gplus\\
     \hline
     Number of vertices & 81,306 & 40,000 & 107,614\\ 
     \hline
     Number of edges & 176,8149 & 20,000,000 & 13,673,453\\
     \hline
     Max outdegree & 3,383 & 1,100 & 17,055\\
     \hline
    \end{tabular}
\end{table}

We take an agent-based modelling approach similar to \cite{brudermann}\cite{abms-influence} and analyse the effectiveness of the model in simulating advertizement campaign dynamics instead of the space or time complexity of the algorithm itself. We make use of two standard metrics to evaluate and compare influencer performance:\\
 
    \textbf{customer acquisition cost ($\xi$) : } It is the cost of winning a customer to purchase a product and is computed as shown in Equation \eqref{eq:cac}, as the ratio of the cost incurred in employing influencers for the advertizing campaign to the number of customers that were obtained as a result of the campaign.
    \begin{equation}
        \xi = \frac{\eta}{\psi}
        \label{eq:cac}
    \end{equation}
    Where $\xi$ is the customer acquisition cost, $\eta$ is the cost of hiring the influencers and $\psi$ is the total number of customers.  \\
    
    \textbf{conversion ratio ($\Theta$) : } It represents the capacity of the influencers to convert potential customers into actual ones. It is computed as shown in Equation \eqref{eq:cr}, as the ratio of the number of customers to the total number of individuals who have participated in the campaign i.e., who have viewed the advertizement.  
    \begin{equation}
        \Theta = \frac{\psi}{\chi}  
        \label{eq:cr}
    \end{equation}
    Where $\Theta$ is the conversion ratio and $\chi$ is the total number of agents which the campaign has reached.

These two metrics are often used in the influencer marketing domain to evaluate the performance of an influencer. Ideally, a marketer would like to acquire customers at as low a cost as possible since this would directly result in a higher return on investment. Hence, a set of influencers with low customer acquisition cost is preferable. Additionally, the conversion ratio gives a good idea of how well an influencer can convince its followers to be a customer. An influencer who might have many followers with a weak influence over them cannot effectively convert followers into customers. It is thus sometimes useful to also evaluate influencers in terms of their ability to convert potential customers. The focus being on a set of influencers, the conversion ratio and the customer acquisition cost are calculated for a set of influencers of a particular kind over an individual influencer. This enables comparison among groups of different kinds of influencers.

Sections \ref{sec:model_validation}, \ref{sec:situational}, and \ref{sec:sweep} present results of experiments on the Twitter social graph. We then experiment on the synthetically generated small world graph in Section \ref{sec:synthetic} to analyze the results on a non-social graph. 

For all the experiments, the following parameters are kept constant: $\sigma=0.2$, $a=0.9$, $\gamma=0.01$ and $c=0.7$. The above values were chosen for these parameters in order to simulate a general circumstance. $\sigma=0.2$ was chosen in order to not let a large number of samples cluster around the mean, i.e., to incorporate diversity in consumer interest. The $a=0.9$ ensured that around 90\% of the network is active \cite{romero-passivity}, and the $\gamma=0.01$ was chosen to maintain a significantly lower spread of negative influence \cite{donny-2018}. Assuming influencer interaction with followers generates a large influence impact on the follower \cite{richardson}, we use the influence update $c = 0.7$. 

The Mesa library in Python was used to design the model and run experiments. Each experimental result was found by taking an average of ten trials.  For each experiment, the graph was initialized once using the initialization parameters and due to the enormity of the studied graph datasets with the number of edges in the order of millions, multiple replications of the same initialization parameters did not alter the conclusions and findings which was observed during simulational testing. The variability of the model in terms of variance is presented to establish the robustness of results and conclusions. 

\subsection{Model Validation}
\label{sec:model_validation}
\begin{figure}[!t]
\centering
\subfloat[Individual performance\label{fig:basic_twitter}]{\includegraphics[width=3in]{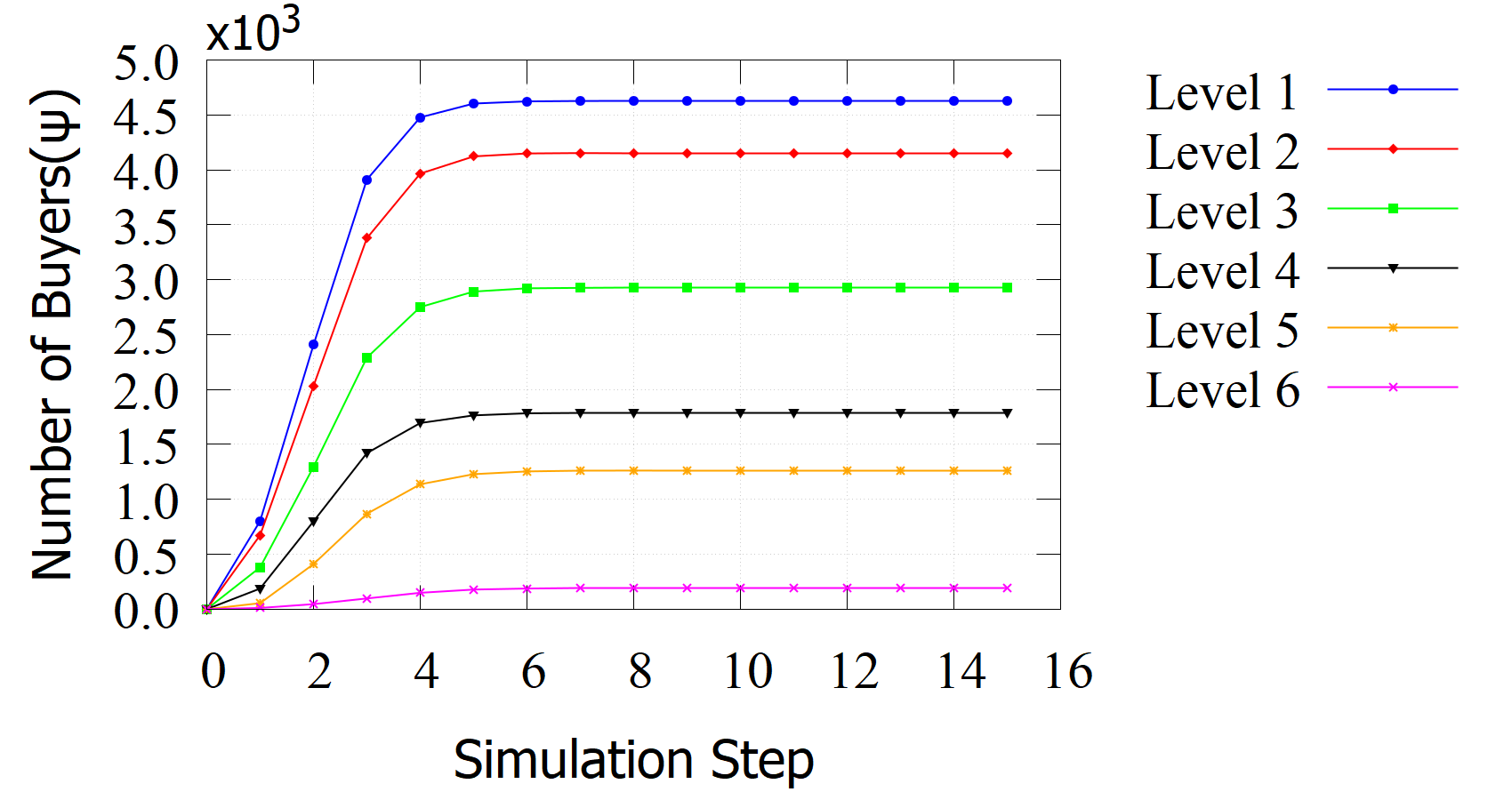}}\qquad
\hfil
\subfloat[Performance of set of influencers\label{fig:twitter_hiring}]{\includegraphics[width=3in]{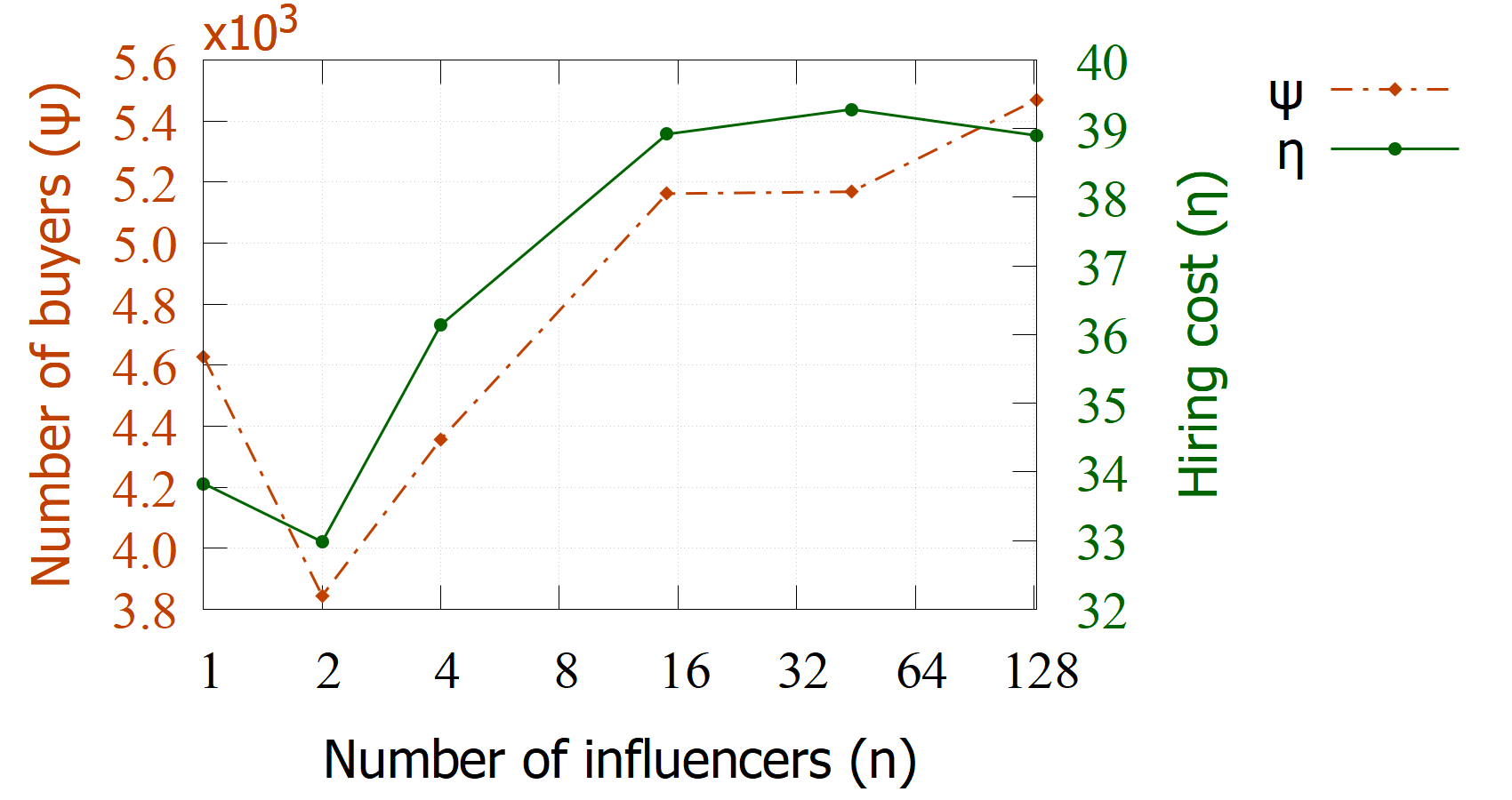}}\qquad
\caption{Validation of individual and set performance of the influencers}
\label{fig:model_val}
\end{figure}

The performance of influencers individually and in sets was evaluated and compared with real-world observations to validate the model. An influencer was chosen at random from each category to begin the advertizing campaign. The agent population was initialized to a medium level of interest: $\lambda_{x} \sim N(\mu=0.5, \sigma=0.2)$, and a medium fraction of willingness: $\Omega=0.5$. The simulation results are shown in Figure \ref{fig:basic_twitter}. In terms of individual performance, it can be observed that a big influencer outperforms a small influencer. A big influencer has a large number of followers, thanks to which its advertizing campaign reaches far and deep into the network. On the other hand, a small influencer has fewer followers and eventually cannot reach out to as many individuals as a big influencer. As a result, the number of customers obtained is significantly lesser compared to a bigger influencer. 

However, brand marketers generally hire multiple influencers across a social network. The performance of a set of influencers and the cost of hiring them varies according to the nature and size of the set. Thus, the next step is to analyze the performance of different kinds of influencers in a set. A set of size $n$ was formed for each influencer type such that the total number of unique followers of the influencers in the set is constant and equal to the number of followers of the biggest influencer in the network, which is $3383$ followers (in the Twitter dataset). This helps to compare the performance of sets of different types of influencers even when their total unique follower count is the same. Since the unique follower count is kept constant, the larger set will comprise of smaller type influencers. The resulting number of buyers and the cost of hiring such a set of influencers is shown in Figure \ref{fig:twitter_hiring}. 

We find that a set of many small influencers attracts more customers than a set of a few big influencers. This is primarily because those small influencers engage with their followers very actively, thereby increasing their influence over them. This engagement is captured by the engagement rate ($\epsilon$) parameter of the model. At the same time, hiring numerous small influencers is found to be more expensive, as indicated by the increasing trend in the hiring cost graph in Figure \ref{fig:twitter_hiring}. Every advertizing campaign tends to have a limited budget and it is therefore unrealistic to hire a great many number of small influencers. A marketer would have to resolve this trade-off between the size of a set and the cost of hiring them according to the end goals and the circumstances in which the campaign will be conducted.   

The results obtained above are fairly intuitive and obey conventional wisdom. The model effectively incorporates consumer and influencer behavior in real-world social network advertizing campaigns. It can thus be used for studying various circumstances and evaluating different strategies.

\subsection{Situational Experiments}
\label{sec:situational}
\begin{figure*}[!t]
\centering
\subfloat[Non-luxury product, $\Omega=0.9$, $\mu=0.5$\label{fig:twitter_w90_i0.5}]{\includegraphics[width=3in]{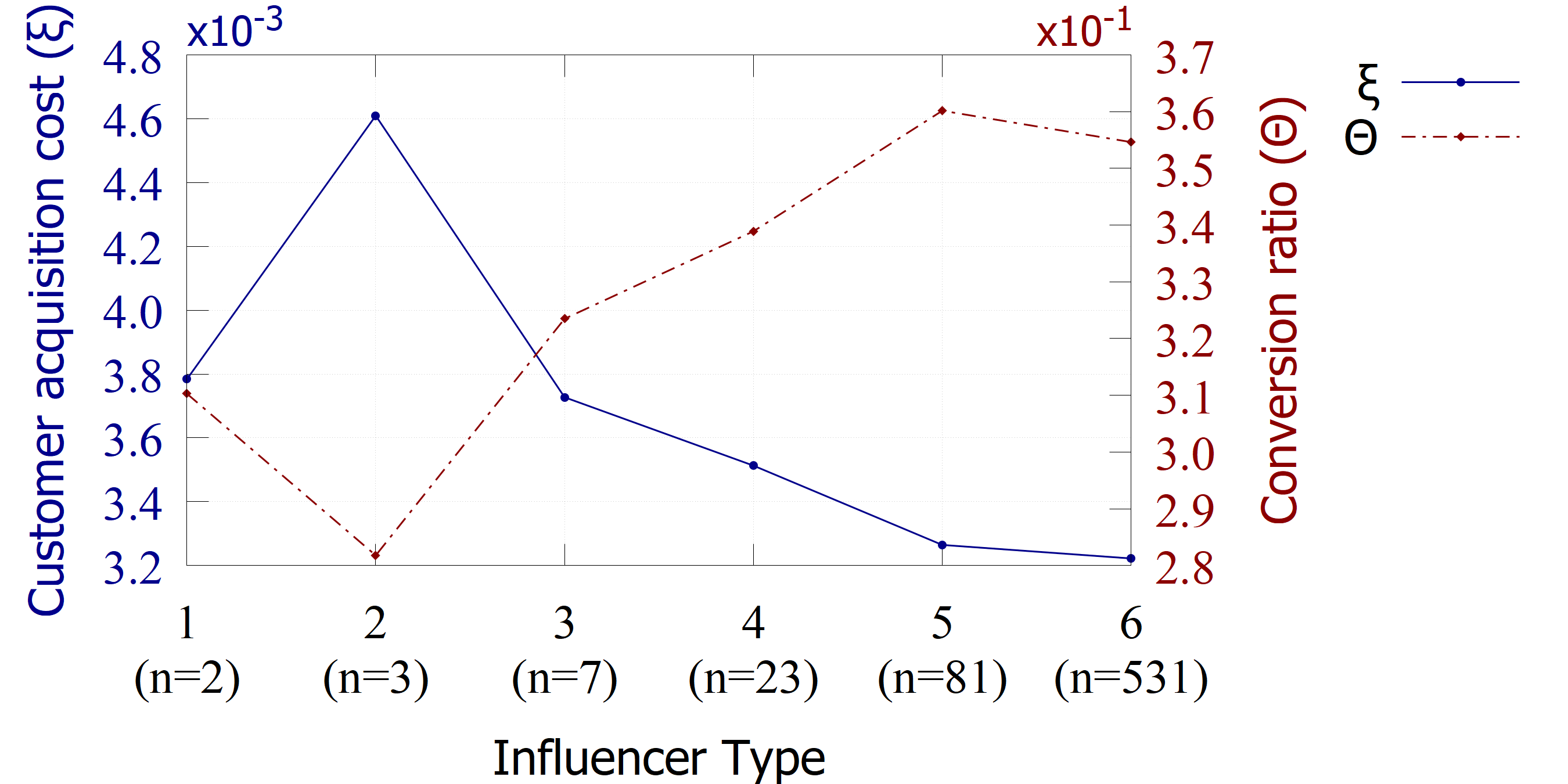}}\qquad
\hfil
\subfloat[Non-luxury product, $\Omega=0.9$, $\mu=0.2$\label{fig:twitter_w90_i0.2}]{\includegraphics[width=3in]{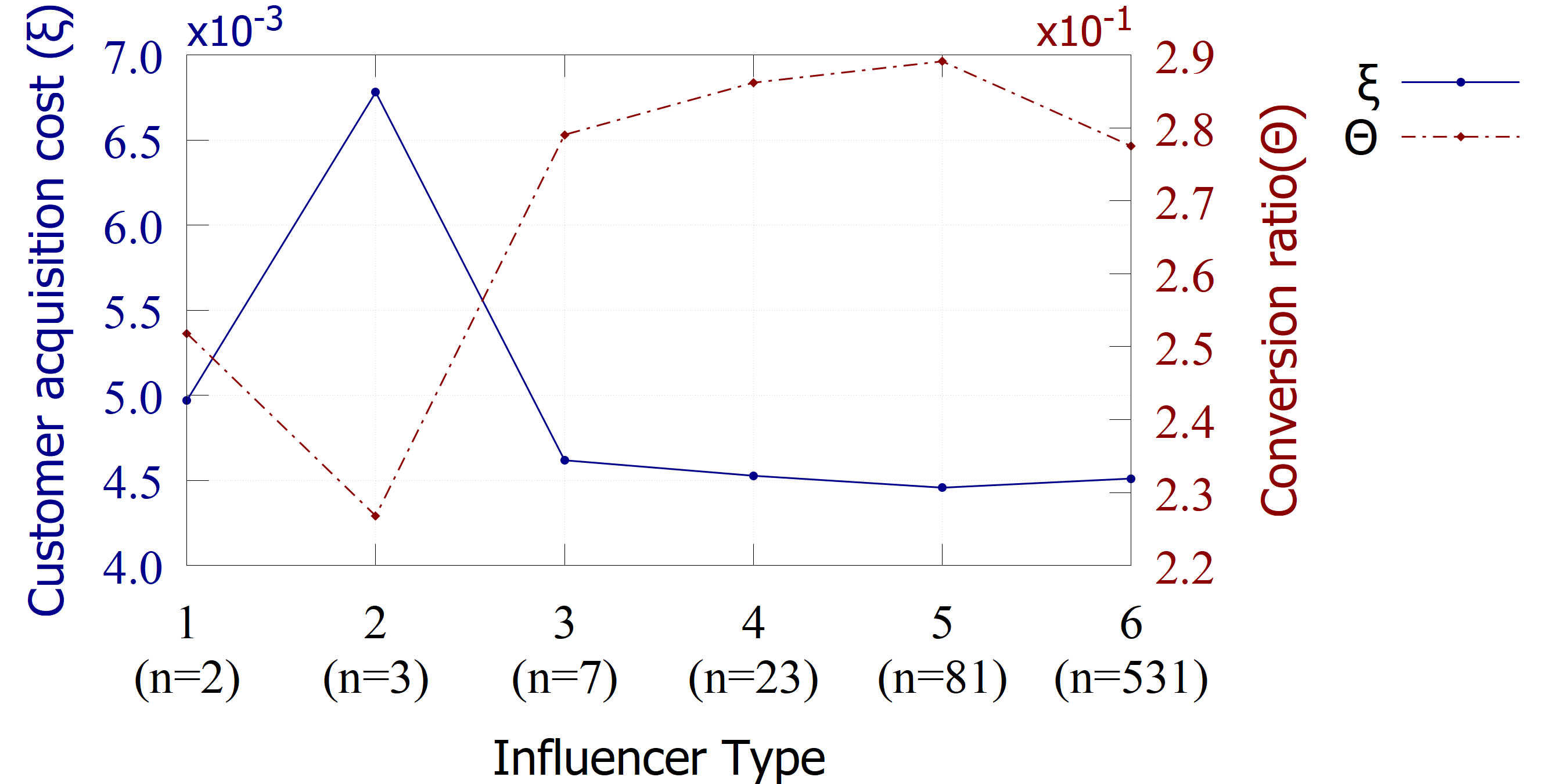}}\qquad
\caption{Simulation results for a non-luxury product with respect to different types of influencer sets, and their corresponding size $n$. The horizontal axis represents categorical data. The customer acquisition cost is found to decrease when using smaller types of influencers whereas the conversion ratio increases.}
\label{fig:results_nonluxury}
\end{figure*}

\begin{figure*}[!t]
\centering
\subfloat[Luxury product, $\Omega=0.1$, $\mu=0.5$\label{fig:twitter_w10_i0.5}]{\includegraphics[width=3in]{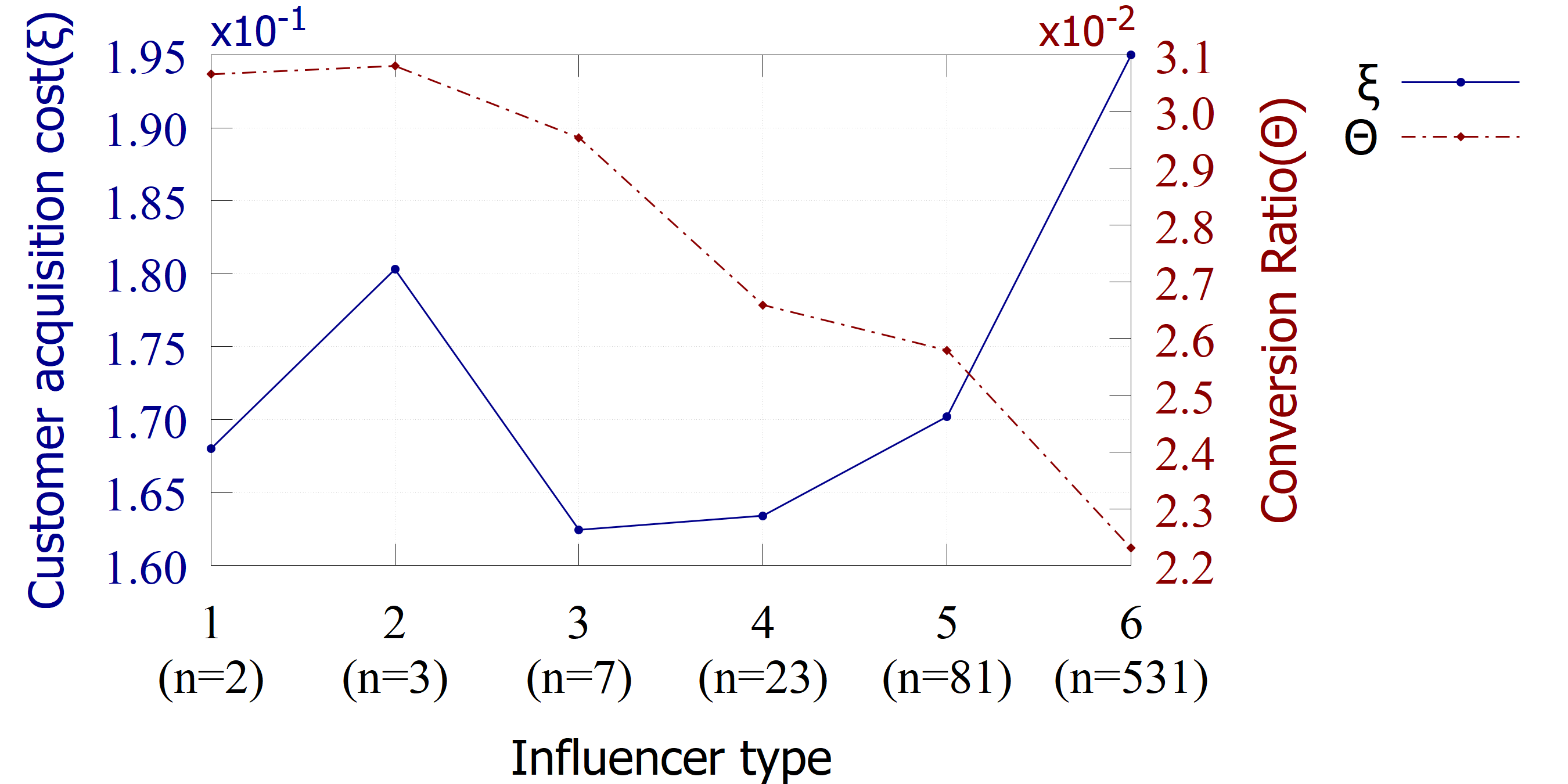}}\qquad
\hfil
\subfloat[Luxury product, $\Omega=0.1$, $\mu=0.8$\label{fig:twitter_w10_i0.8}]{\includegraphics[width=3in]{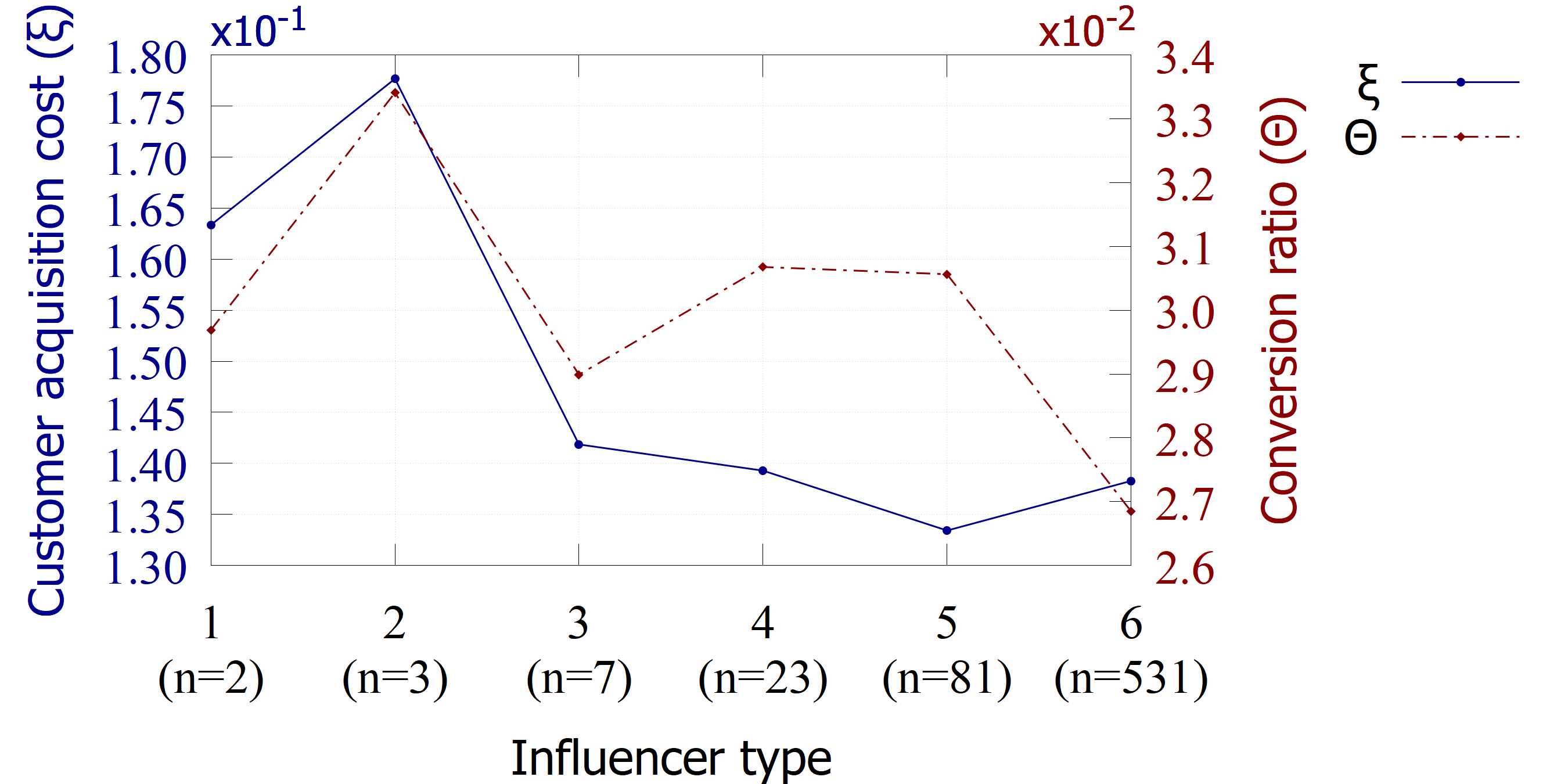}}\qquad
\caption{Simulation results for a luxury product with respect to different types of influencer sets, and their corresponding size $n$. The horizontal axis represents categorical data. (a) The customer acquisition cost is found to increase when using smaller types of influencers whereas the conversion ratio decreases. (b) When the customers' interest in the product is already high, both the customer acquisition cost and the conversion ratio decrease when using smaller types of influencers.}
\label{fig:results_luxury}
\end{figure*}
The performance and subsequent choice of influencers are dependent on the marketing situation, of which the nature of the product to be marketed is a major factor. We consider two kinds of products for experimentation\textemdash non-luxury and luxury products.

We simulate the advertizing campaign for the two kinds of products mentioned above. We make a reasonable assumption that the campaign has a limited budget to hire influencers. We set the hiring investment limit $\rho = 68$. We choose this value since the biggest influencer in the Twitter social graph has $3383$ followers with a hiring cost of $\$33.83$, and hence, $\rho = 68$ is sufficient to hire at least two celebrities to facilitate a reasonable comparison with other types of influencers. 

A set of influencers of each type is hired according to the investment limit $\rho$. For each set, $n$ advertizers are chosen as shown in Equation \ref{hiring_eq}.
\begin{equation}
    \eta = \sum_{i=1}^{n}h_{i}, \ \text{such that} \ \eta = \rho
    \label{hiring_eq}
\end{equation}
Where $\eta$ is the total hiring cost of the set and $h_{i}$ is the cost of hiring influencer agent $i$. Since smaller type influencers are less expensive to hire, larger sets can be formed from them at the same budget, as compared to a set of big influencers. 

\subsubsection{Non-luxury product}
\label{non-luxury}
An advertizing campaign of a non-luxury product is simulated by setting the fraction of network willing to pay $\Omega$ to a high value, which means that most of the population is willing to pay for the product. In this case we set $\Omega=0.9$. We present two cases: when the initial customers' interest in the product is medium, $\lambda \sim N\left(\mu=0.5,\sigma=0.2\right)$ which is shown in Figure \ref{fig:twitter_w90_i0.5}, and when the initial interest is low, $\lambda \sim N\left(\mu=0.2,\sigma=0.2\right)$ which is shown in Figure \ref{fig:twitter_w90_i0.2}. The customer acquisition cost and the conversion ratio are computed from the simulation results and are shown in Figure \ref{fig:results_nonluxury}.

The following observations can be drawn from Figure \ref{fig:results_nonluxury}:
\begin{itemize}
    \item The conversion ratio from celebrities (type 1) to nano-influencers (type 6) is roughly increasing in both Figure \ref{fig:twitter_w90_i0.5} and \ref{fig:twitter_w90_i0.2}. This means that nano and micro-influencers are better at converting their sub-networks into customers than mega-influencers and celebrities.
    \item The customer acquisition cost is roughly decreasing in both Figure \ref{fig:twitter_w90_i0.5} and \ref{fig:twitter_w90_i0.2}. This means that it is less expensive to acquire customers using nano-influencers than celebrities.
\end{itemize}

We observe some dips or spikes in the customer acquisition cost or the conversion ratio (e.g., for influencer type 2). These are due to the nature of the graph and the particular influencer agent/vertex. However, the general observation that the conversion ratio increases from celebrities to nano-influencers and that the customer acquisition cost decreases, still holds. Similar trends are observed from experimentation on the Gplus social graph which is discussed later in Appendix \ref{apdx:appendix}. 

Considering celebrities and nano-influencers, the above observations allow us to conclude that for non-luxury products with a medium ($\mu=0.5$) or low ($\mu=0.2$) level of initial customers' interest, hiring nano-influencers is the better influencer marketing strategy. 

\subsubsection{Luxury product}
\label{luxury}
A advertizing campaign of a luxury product is simulated by setting the fraction of willingness $\Omega$ parameter to a low value, which means that most of the population is unwilling to pay for the product. In this case we set $\Omega=0.1$. We present two cases: when the initial customers' interest in the product is medium, $\lambda \sim N\left(\mu=0.5,\sigma=0.2\right)$, and when the initial interest is high, $\lambda \sim N\left(\mu=0.8,\sigma=0.2\right)$. It is important to note that the interest in a product can be independent of the willingness to pay. For example, many people might be interested in and fascinated by a luxury car, but not all of them intend to buy one. 

The customer acquisition cost and the conversion ratio are computed from the simulation results and shown in Figure \ref{fig:results_luxury}. 

The following observations can be drawn from Figure \ref{fig:twitter_w10_i0.5}:
\begin{itemize}
    \item The conversion ratio is decreasing meaning that nano and micro-influencers are worse at converting their sub-networks into customers than mega-influencers and celebrities.
    \item The customer acquisition cost is roughly increasing meaning that it is less expensive to acquire customers using celebrities than nano-influencers.
\end{itemize}

The following observations can be drawn from Figure \ref{fig:twitter_w10_i0.8}:
\begin{itemize}
    \item The conversion ratio is decreasing meaning that nano and micro-influencers are worse at converting their sub-networks into customers than mega-influencers and celebrities.
    \item The customer acquisition cost is roughly decreasing meaning that it is less expensive to acquire customers using nano-influencers than celebrities.
\end{itemize}

\begin{figure*}[!t]
\centering
\subfloat[Sweep with respect to the customer acquisition cost\label{fig:twitter_sweep_cac}]{\includegraphics[width=3in]{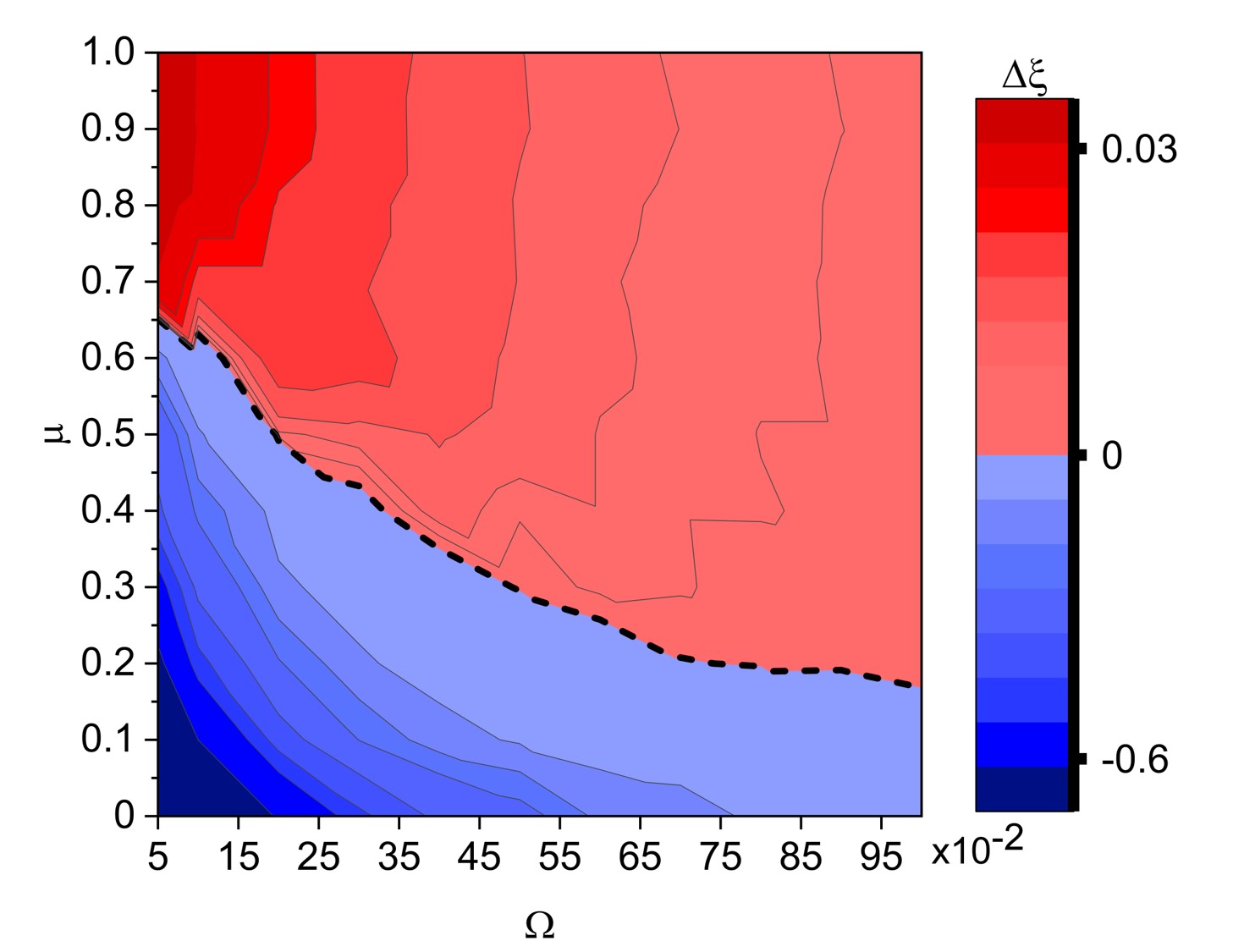}}\qquad
\hfil
\subfloat[Sweep with respect to the conversion ratio\label{fig:twitter_sweep_cr}]{\includegraphics[width=3in]{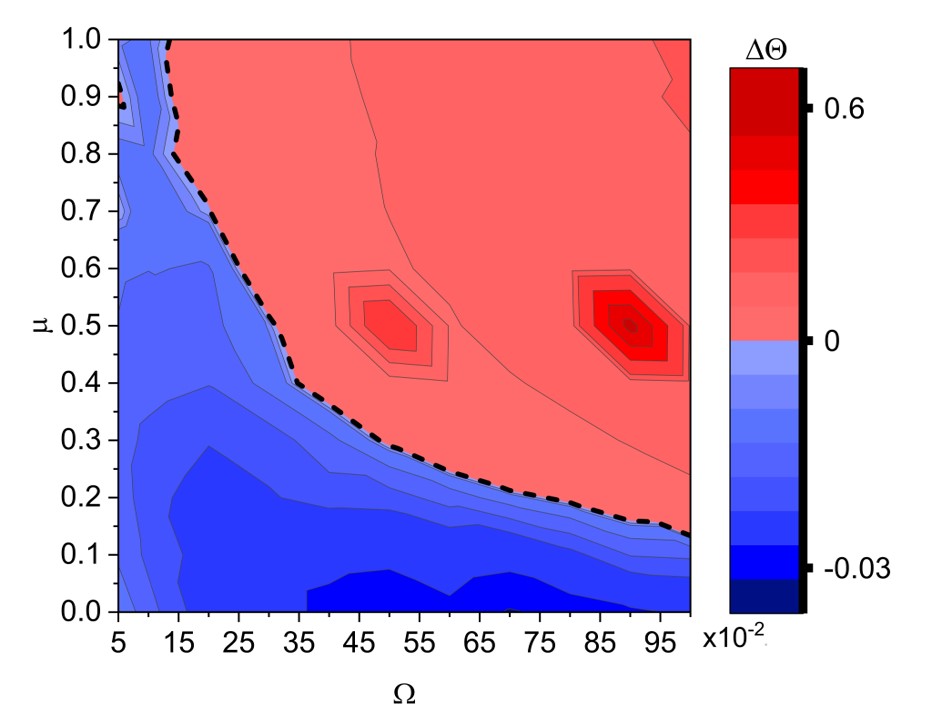}}\qquad
\caption{Results of the parametric sweep for the Twitter graph represented as a contour map: the dashed line represents the boundary between celebrities (in blue) or nano-influencers (in red) being the better performers. The gradient represents the magnitude of their difference.}
\label{fig:results_sweep}
\end{figure*}
Considering celebrities and nano-influencers, the above observations allow us to conclude that for luxury products with a medium level of initial customers' interest ($\mu=0.5$), hiring celebrities is the better influencer marketing strategy. However, if the initial customers' interest is high ($\mu=0.8$), hiring nano-influencers would be the better strategy. This might be because a large proportion of the population already has a high degree of interest and this reduces the role that influencers have to play to increase the interest of the population. Owing to this high degree of interest, the campaign spreads significantly deeper and nano-influencers perform better in terms of the customer acquisition cost even though their conversion ratio remains lower than the celebrities. This is an important result because it shows the significance of the customers' interest in the product on the outcome of advertizing campaigns. 

Overall, we find from Figure \ref{fig:results_nonluxury} and \ref{fig:results_luxury} that the order of magnitude of the customer acquisition cost for a luxury product is much higher than for a non-luxury product. In other words, it is thus more expensive to acquire customers when marketing a luxury good as compared to marketing a non-luxury good. 

\begin{table}[h!]
  \centering
    \caption{Variability of the situational experiments}
    \label{tab:results_var}
    \begin{tabular}{|c|c|c|}
    \hline
      & \multicolumn{2}{|c|}{\textbf{Variance}}\\
     \hline
     \textbf{Experiment} & $\xi$ & $\Theta$ \\
     \hline
     Non-luxury product, $\Omega=0.9$, $\mu=0.5$ & 3.69e-7 & 1e-3\\
     \hline
     Non-luxury product, $\Omega=0.9$, $\mu=0.2$ & 1.03e-6 & 6.58e-4\\
     \hline
     Luxury product, $\Omega=0.1$, $\mu=0.5$ & 4.61e-4 & 2.05e-5\\
     \hline
     Luxury product, $\Omega=0.1$, $\mu=0.8$ & 4.54e-4 & 1.17e-5\\
     \hline
    \end{tabular}
\end{table}

The variability of these experiments in terms of the variance of customer acquisition cost $\xi$ and conversion ratio $\Theta$ is shown in Table \ref{tab:results_var}. The variance was computed for the ten trials. The low variability indicates the robustness of the model to small random variations. 

\subsection{A Parametric Sweep}
\label{sec:sweep}
The results obtained in subsections \ref{sec:situational} and \ref{sec:model_validation} exhibit the situational nature of advertizing outcomes. The performance of a set of influencers was shown to depend on the customers' interest in the product and the kind of product being advertized. In addition, it also depends on the structure of the social network itself. It is hence necessary to obtain an overall picture of how an advertizing campaign propagates in various scenarios. We do this by performing a parametric sweep, in simulation, of the mean of customers' interest $\mu$ and their willingness to pay $\Omega$. Recall that the willingness to pay captures the nature of the marketed product i.e., low values represent a luxury product and high values represent a non-luxury product. The two parameters are varied as shown below:
\begin{itemize}
    \item The mean is varied as, $0 < \mu \leq 1$, in steps of $0.1$; $\mu=0$ is excluded because the simulation would result in no buyers.
    \item The fraction of willingness is varied as $0.05 \leq \Omega \leq 1$. We start at $\Omega=0.05$ because it was found that lower values result in no customers for this particular graph.
\end{itemize}
The influencers are hired similar to previous experiments, i.e., as described by Equation \eqref{hiring_eq}. The parameteric sweep is performed while computing the difference in the customer acquisition cost of the set of celebrities ($\xi_{1}$) and nano-influencers ($\xi_{6}$). The difference of the conversion ratio of the set of celebrities ($\Theta_{1}$) and nano-influencers ($\Theta_{6}$) is also computed during the sweep.  
\begin{itemize}
    \item Difference in customer acquisition cost:
    $\Delta \xi =\xi_{1} - \xi_{6}$
    \item Difference in conversion ratio: 
    $\Delta \Theta = \Theta_{1} - \Theta_{6}$
\end{itemize}

The results of the parametric sweep are represented as a contour map of the two parameters as shown in Figure \ref{fig:results_sweep}. Similar to the previous experiments, the type of influencer with a lower customer acquisition cost is considered the better investment. The dashed contour line indicates the boundary between the the situations in which the nano-influencers and celebrities are to be preferred. The regions in which celebrities are the recommended choice are denoted in blue and the regions in which the nano-influencers are the recommended choice are denoted in red.

Considering the customer acquisition cost, we find from from Figure \ref{fig:twitter_sweep_cac} that:
\begin{itemize}
    \item Nano-influencers are a better investment when advertizing any kind product (luxury or non-luxury) when the mean customers' interest in the product is higher than $0.65$, i.e., $\mu > 0.65$.
    \item On the other hand, celebrities are the better investment when advertizing any kind product when the mean interest is low: approximately less than $0.2$, i.e., $\mu < 0.2$.
    \item However, for a medium level of customers' interest ($\mu=0.5$), the general trend is that celebrities are a better investment to market luxury products and nano-influencers become increasingly better for marketing non-luxury products, i.e., as the fraction of willingness $\Omega$ increases.
\end{itemize}

\begin{figure*}[!t]
\centering
\subfloat[Sweep with respect to the customer acquisition cost\label{fig:syn_sweep_cac}]{\includegraphics[width=3in]{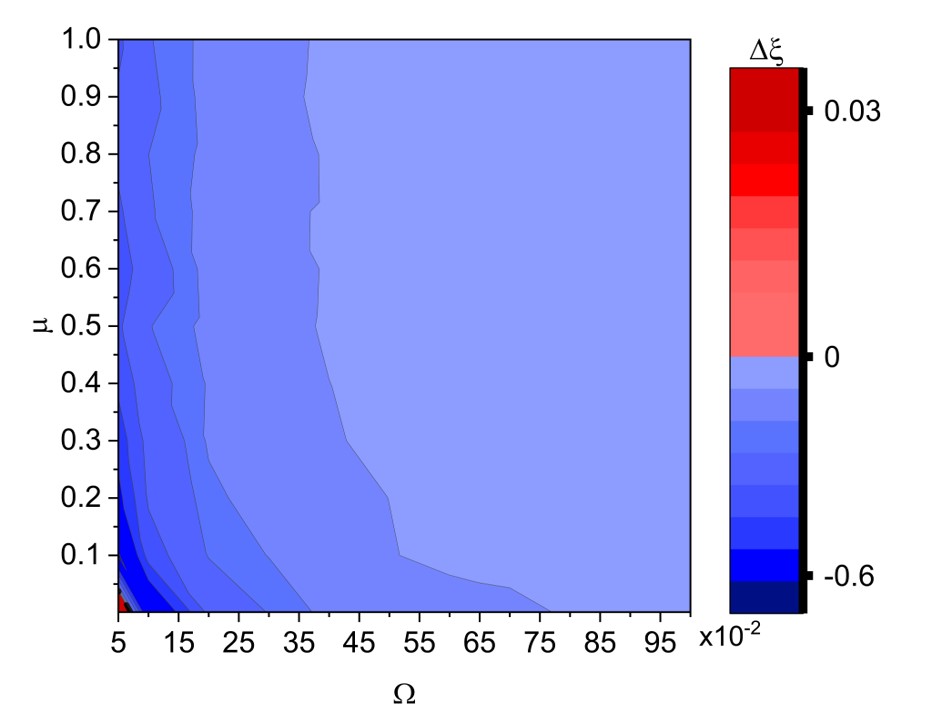}}\qquad
\hfil
\subfloat[Sweep with respect to the conversion ratio\label{fig:syn_sweep_cr}]{\includegraphics[width=3in]{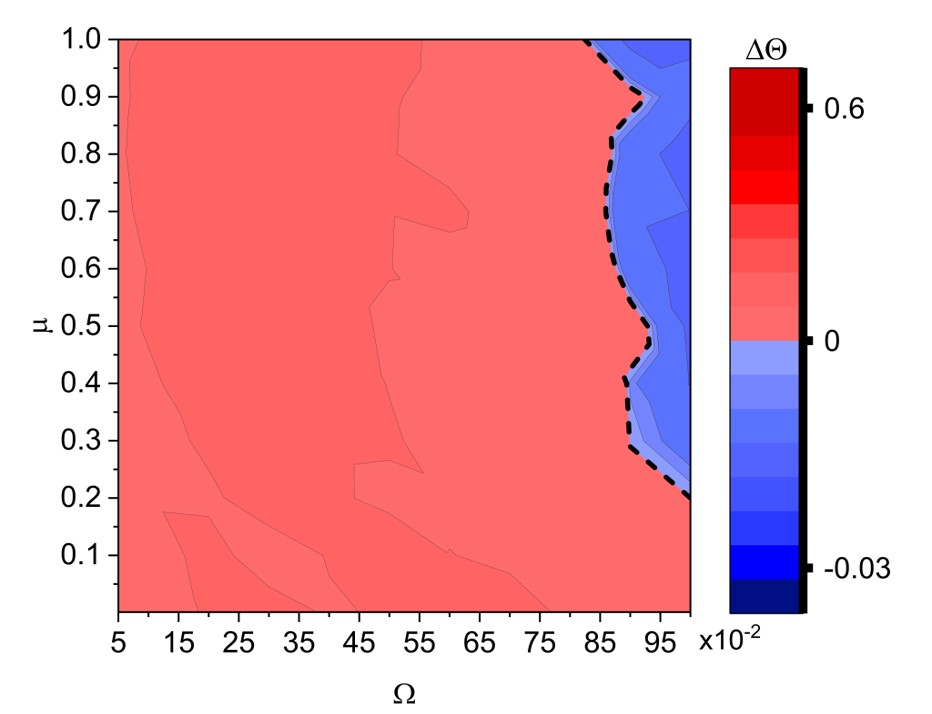}}\qquad
\caption{Results of the parametric sweep for the synthetic graph represented as a contour map: the dashed line represents the boundary between celebrities (in blue) or nano-influencers (in red) being the better performers. The gradient represents the magnitude of their difference.}
\label{fig:results_syn_sweep}
\end{figure*}

Considering the conversion ratio, shown in Figure \ref{fig:twitter_sweep_cr}, we find that:
\begin{itemize}
    \item Celebrities are better at converting potential customers when $0.05 < \Omega < 0.15$, irrespective of the mean custmers' interest in the product.
    \item Irrespective of the kind of product, for a very low interest, i.e., $\mu<0.1$, celebrities outperform nano-influencers in terms of conversion. 
\end{itemize}

Interestingly, the regions of performance in terms of the conversion ratio (Figure \ref{fig:twitter_sweep_cr}) and customer acquisition cost (Figure \ref{fig:twitter_sweep_cac}) do not coincide. This can be observed for low values of $\Omega$ and high values of $\mu$. In other words, a higher rate of converting followers and their sub-networks into customers does not always lead to cheaper acquisition of customers. For instance, consider an influencer $x_{1}$ whose campaign has reached $\chi_{1}$ individuals with a conversion ratio $\Theta_{1}$, and acquired $\psi_{1}$ customers at a cost of $\xi_{1}$.
Consider another influencer $x_{2}$ whose campaign has reached $\chi_{2}$ individuals with a conversion ration $\Theta_{2}$, and acquired $\psi_{2}$ customers at a cost of $\xi_{2}$. Suppose, $\Theta_{1} > \Theta_{2}$ but $\chi_{1} < \chi_{2}$. This means that the campaign by $x_{1}$ reached fewer individuals in total, but converted a higher fraction of them as compared to $x_{2}$. Yet, since $x_{2}$ has reached a larger number of individuals in the network, it is possible that the total number of customers acquired by $x_{2}$ is higher than those acquired by $x_{1}$, i.e., $\psi_{2} > \psi_{1}$, where $\psi_1$ and $\psi_2$ are computed from Equation \eqref{eq:cr} as $\psi_1 = \Theta_1 \cdot \chi_1$ and similarly for $\psi_2$.

It follows from the inverse relation between the hiring cost $\eta$ and the customer acquisition cost $\xi$, indicated in Equation \eqref{eq:cac} that for the same hiring cost, $\xi_{1} > \xi_{2}$. Thus, although influencer $x_{1}$ has a higher conversion ratio, it has a higher customer acquisition cost. Hence, the customer acquisition cost does not depend only on the conversion ratio, but also on the total number of agents that an influencer can reach (its outreach).    

\subsection{Results on a Synthetic Graph}
\label{sec:synthetic}

The experiments and results presented up to now were conducted on the twitter graph dataset. As already indicated, the outcome is also dependent on the nature of the graph. To show how the model works on a different kind of graph, we also experiment on a synthetic small world graph generated using the Watts Strogatz model \cite{watts-model}. This model was chosen because social networks are known to possess small-world properties \cite{networks-crowds-book}\cite{exp-study-smallworld}\cite{smallworld}. However, unlike real world social networks, the outdegree distribution of this graph does not obey the power law (Figure \ref{fig:syn_dist}). We perform a parametric sweep just like in Section \ref{sec:sweep} and the contour map of the results are shown in Figure \ref{fig:results_syn_sweep}. These results are quite different from the results of the Twitter dataset, mainly because the outdegree distribution of the synthetically generated graph is different. As a result, observations made on real world social networks do not hold. Here, we find that the celebrity is almost always a better investment since their customer acquisition cost is lower than the nano-influencers for the same hiring investment. This can be seen in Figure \ref{fig:syn_sweep_cac}. However, the conversion ratio of nano-influencers is found to be higher for the most part as seen in Figure \ref{fig:syn_sweep_cr}. 

In conclusion, a parametric sweep of the nature of the product and the customers' interest gives a good overall picture of the outcome of an advertizing campaign for varying situations, with respect to the different types of influencers within the social network. 

\subsection{Discussion}
\label{section_dis}
This model is theory-driven and designed for exploratory studies. We test the model on two datasets\textemdash a Twitter social graph \cite{snap-twitter} and a synthetically generated small-world graph. The correctness of our results is seen by comparison with real-world observations. Each experimental result is found by taking an average of 10 trials. We first perform experiments to validate the correctness of the model in terms of its realistic behavior. Here, our results suggest that an advertizing campaign run individually by an influencer with a greater number of followers generates more sales than a campaign run by an influencer with fewer followers. 

Influencer advertizing campaigns, however, are generally implemented by employing multiple influencers throughout the social network. Thus, we consider situations where a set of influencers are employed to advertize a product, such that the total unique number of followers is the same, and show that the sales increase as the size of the set increases (larger set will consist of smaller influencers). The increase in set size is accompanied by an increase in the hiring cost and hence, a trade-off is involved in choosing a suitable set of influencers. The resolution of this trade-off depends on the circumstances under which the advertizing campaign is conducted. 

The advertizing circumstances we consider are the nature of the product being advertized viz. luxury and non-luxury, and the customers' interest in the product. Other factors that determine the circumstance include the engagement of influencers, the activeness of an individual in the social network, and the structure of the social network. We simulate advertizing campaigns for varying kinds of products and customers' interest and evaluate the performance of different kinds of influencers. We use two well-known metrics to evaluate the performance of influencers\textemdash \emph{customer acquisition cost} \cite{cac} and the \emph{conversion ratio}. Our experiments lead to an important finding\textemdash a higher conversion ratio does not always mean a lower customer acquisition cost and vice-versa. In other words, a higher rate of converting followers into customers does not always lead to the cheaper acquisition of customers.

For a specific hiring investment, the influencers who provide a lower customer acquisition cost are preferable since this directly corresponds to higher gains. In our experiments, we show how the performance of influencers changes as we vary the nature of the product and customers' interest. In the Twitter social graph dataset \cite{snap-twitter}, we find that for a medium level of customers' interest, celebrities are a better investment to advertize luxury products, and nano-influencers are better to advertize non-luxury products. The reasons being that in the former case, celebrities acquire customers at a lower cost than nano-influencers, and in the latter case, it is vice-versa. Additionally, we find that the performance of nano-influencers improves with an increase in customers' interest whereas the performance of celebrities declines. The selection of influencers for advertizing is thus shown to depend on the marketing circumstances. It is due to this that our study provides practical insight which can be very useful for research in the domain of influencer marketing. 

\section{Conclusion}
\label{section_conc}
 
Our model for influencer marketing campaigns has been validated for realistic behavior and used to study the trade-offs between different types of influencers and marketing strategies. From our experiments, we show how the best strategies depend on various parameters and there is no one-point solution for finding the best set of influencers. The results for different parameters were presented in the parametric sweep which can help marketers analyze and improve their influencer marketing strategies.

Our model is designed to be flexible in the sense that inputs that are relevant to the marketing scenario can be provided as parameters for simulation. These inputs are generally obtained by marketers through surveys, questionnaires, etc. In order to simulate a specific advertizing plan/campaign, a social network of choice in the form of a graph, the corresponding customers' interest and willingness to pay, the influencer hiring and engagement rates of the network, etc. can be provided as input for simulation. Hence, our model can also be used as a base for further simulation aided studies in the relatively new domain of influencer marketing. An important direction in which this work can be extended is to understand the mechanisms by which fake news, hate speech, and malicious gossip are spread in social networks through influencers. This in turn
can lead to a better understanding of how myths, rumors, and misunderstandings are propagated by influencers, and can help devise strategies to counter such malicious influencers.

\appendix[Experimental results using the Gplus social graph]
\label{apdx:appendix}

\begin{figure}[!t]
  \centering
  \includegraphics[width=3in]{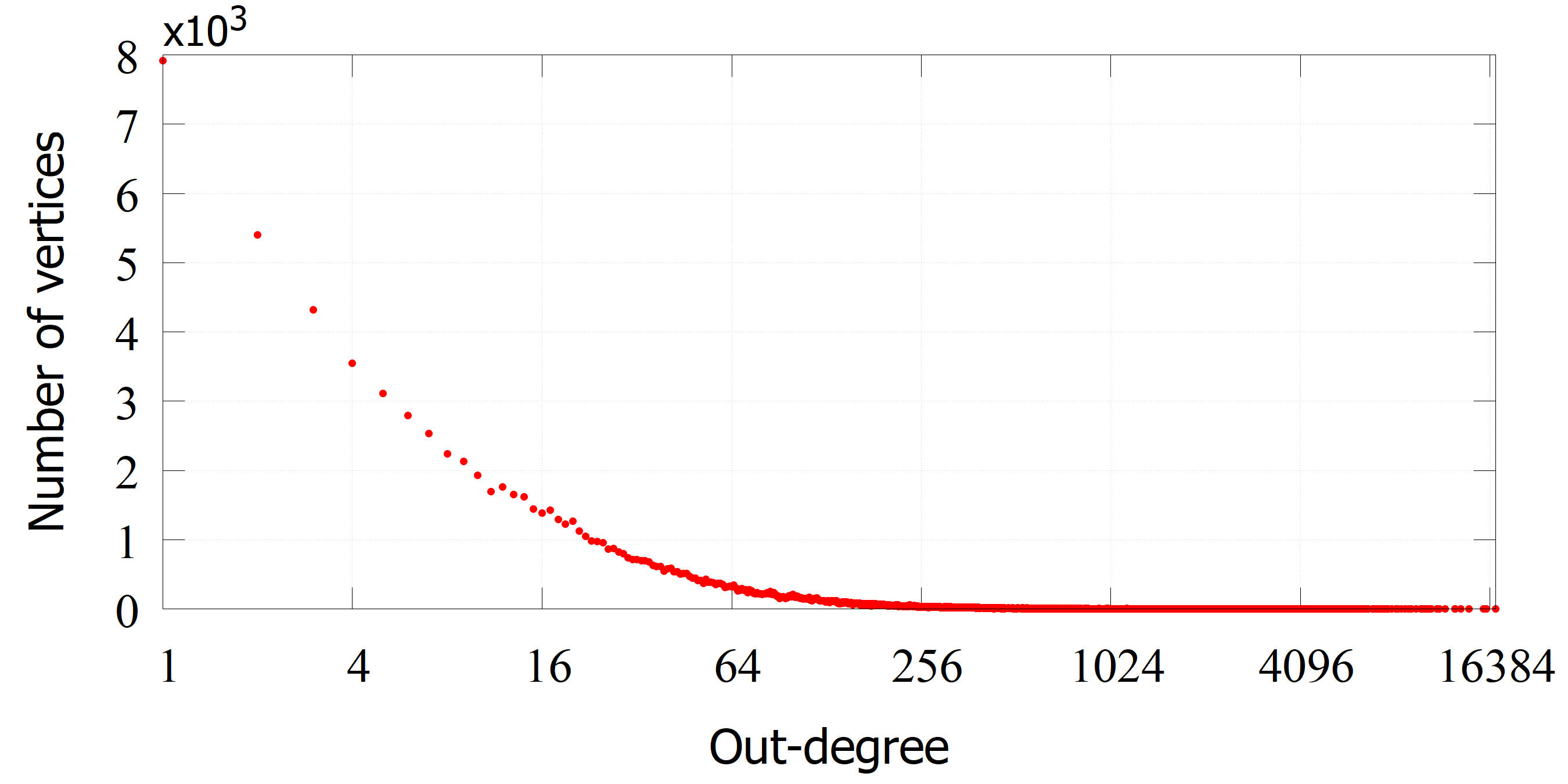}
  \caption{Gplus Outdegree distribution}
  \label{fig:gplus_degree_dist}
\end{figure}

We also perform the experiments discussed in Section \ref{sec:model_validation} and \ref{sec:situational} on another real-world dataset\textemdash the Gplus social graph \cite{snap-twitter}. The out-degree distribution of the Gplus social network is shown in Figure \ref{fig:gplus_degree_dist}. Just like the Twitter social graph, the outdegree distribution obeys the power law. 

The situational experiments as described in Section \ref{sec:situational} where the performance of different types of influencers are compared for luxury and non-luxury products at different mean customers' interest are shown below:
\subsection{Non-luxury}
\begin{figure*}[!t]
\centering
\subfloat[Non-luxury product, $\Omega=0.9$, $\mu=0.5$\label{fig:gplus_w90_i0.5}]{\includegraphics[width=3in]{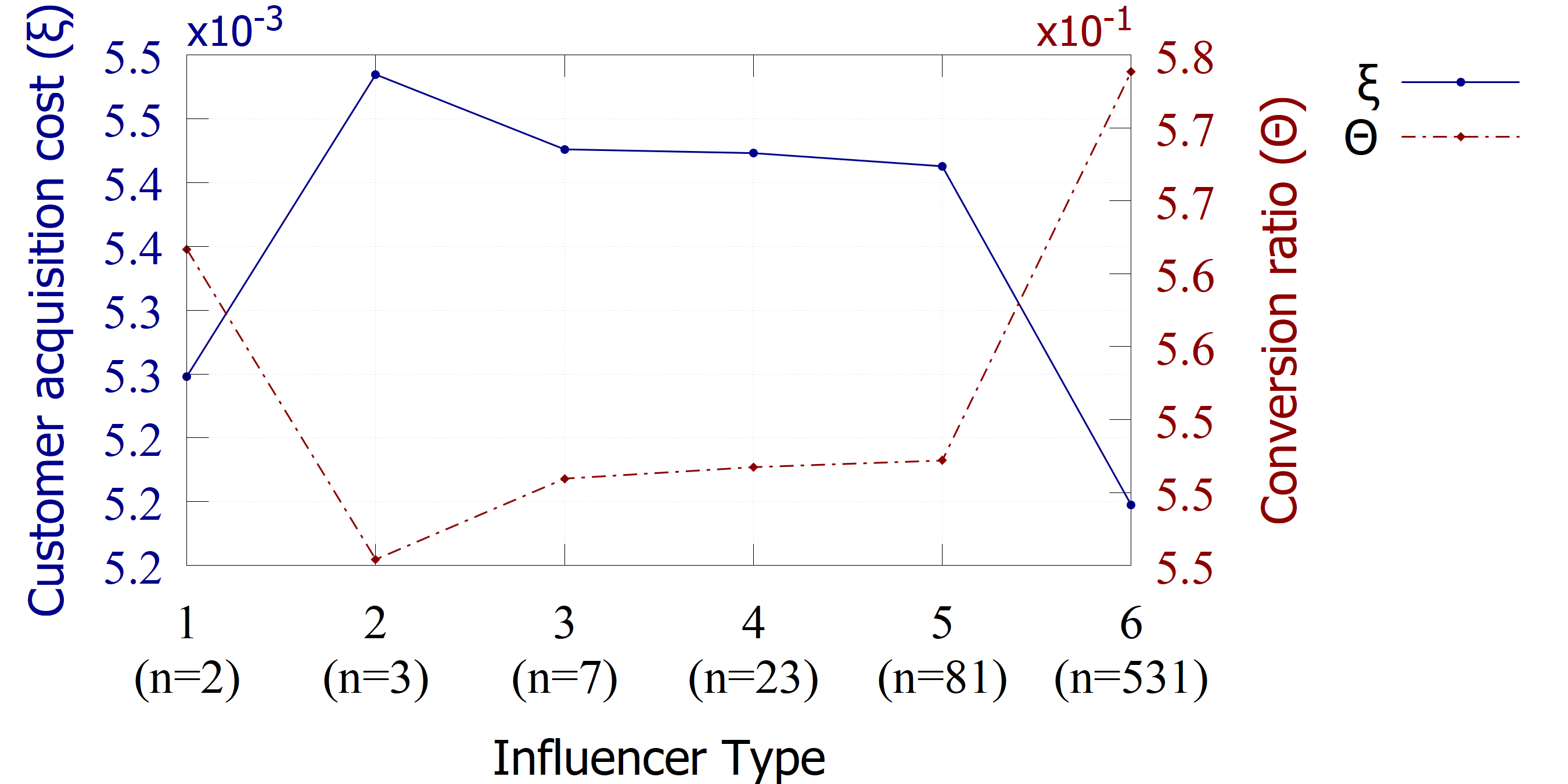}}\qquad
\hfil
\subfloat[Non-luxury product, $\Omega=0.9$, $\mu=0.2$\label{fig:gplus_w90_i0.2}]{\includegraphics[width=3in]{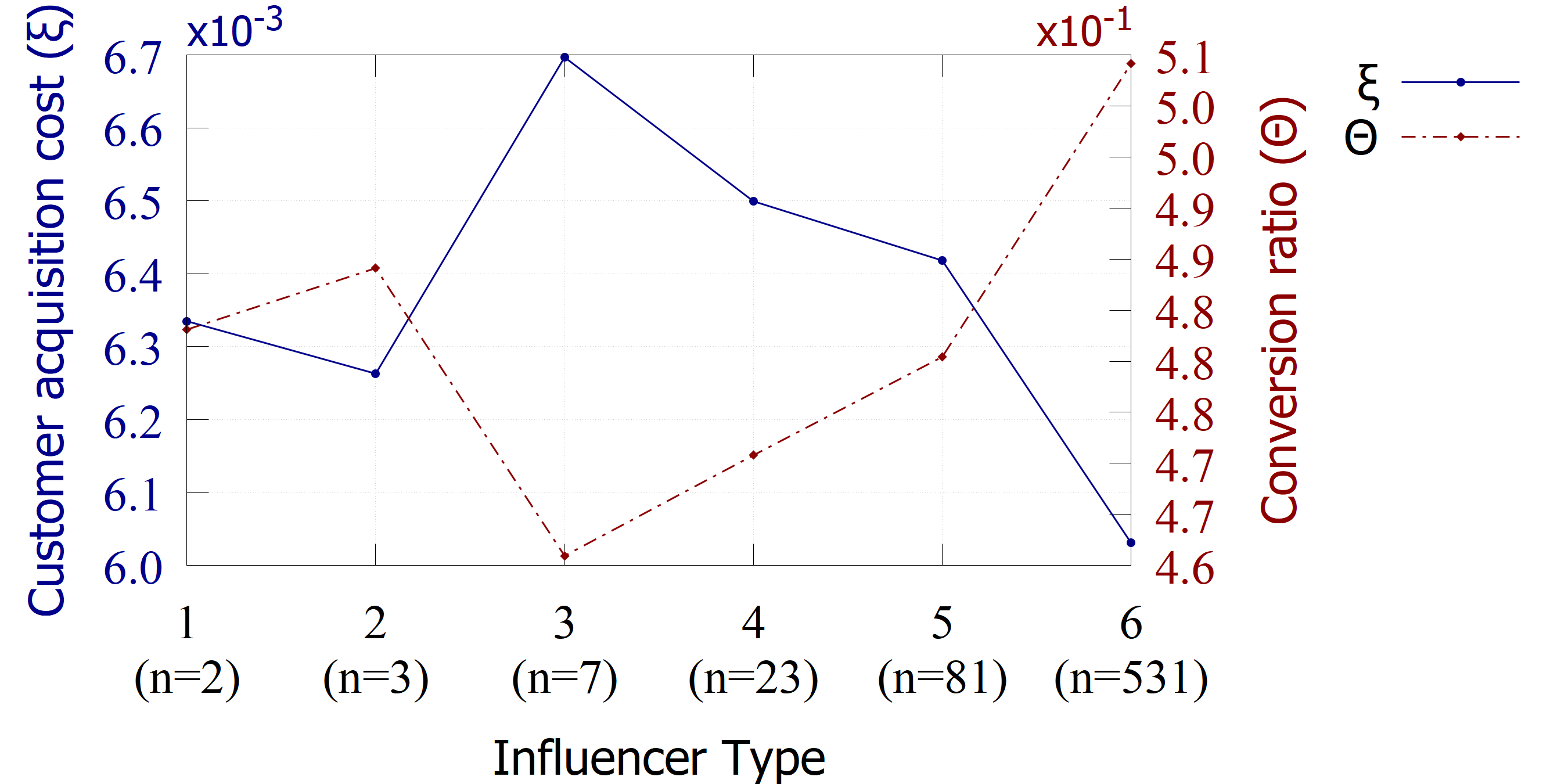}}\qquad
\caption{Simulation results for a non-luxury product with respect to different types of influencers hired according to the budget $\rho$.}
\label{fig:results_nonluxury_gplus}
\end{figure*}

Similar to the observations made on the Twitter graph in Section \ref{non-luxury}, we find that for both cases as shown in Figure \ref{fig:gplus_w90_i0.5} and \ref{fig:gplus_w90_i0.2}: (a) Nano-influencers have a lower customer acquisition cost than celebrities, meaning it is less expensive to acquire customers using nano-influencers. (b) Nano-influencers have a higher conversion ratio that celebrities.

\subsection{Luxury}
\begin{figure*}[!t]
\centering
\subfloat[Luxury product, $\Omega=0.1$, $\mu=0.5$\label{fig:gplus_w10_i0.5}]{\includegraphics[width=3in]{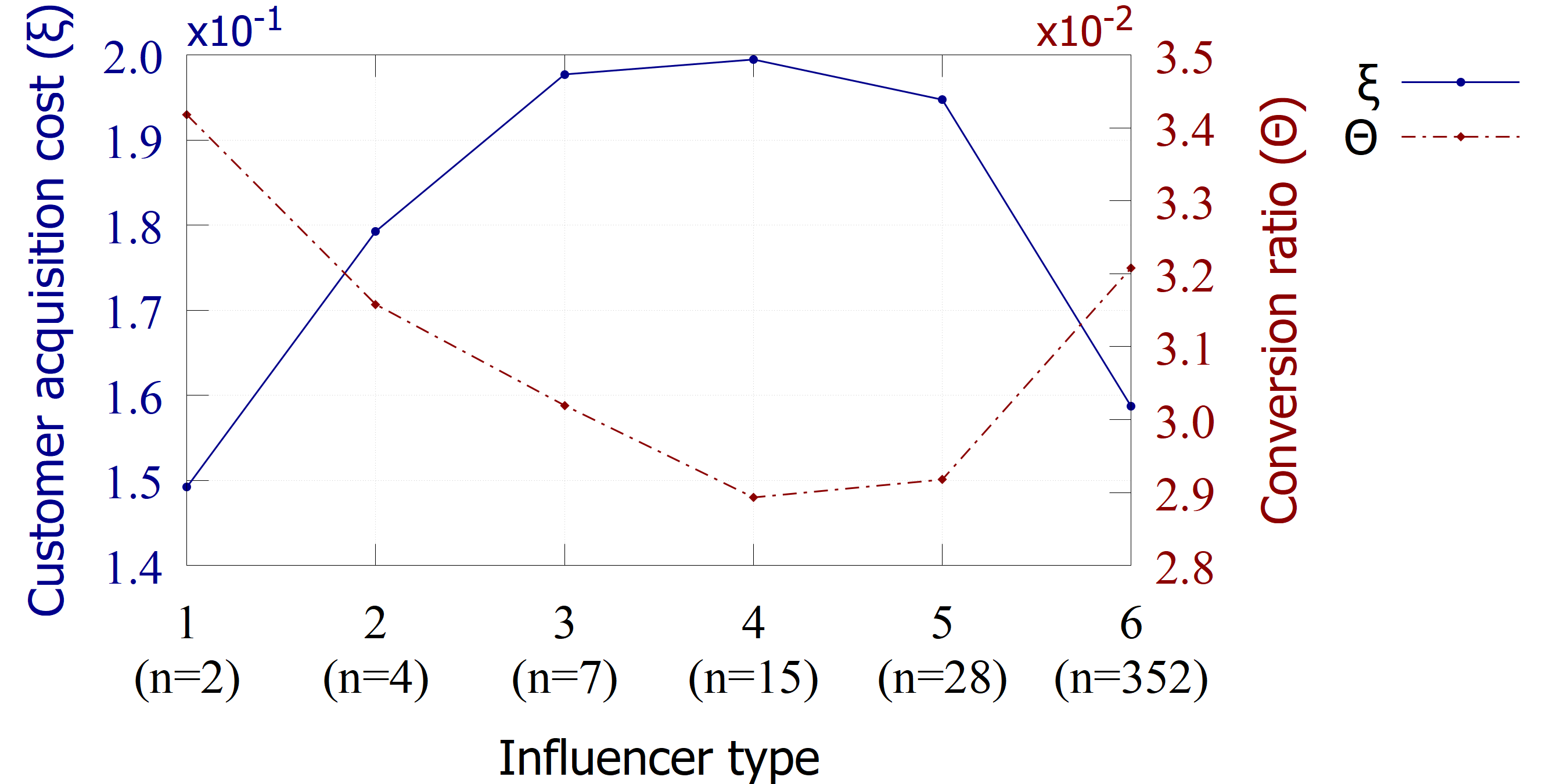}}\qquad
\hfil
\subfloat[Luxury product, $\Omega=0.1$, $\mu=0.8$\label{fig:gplus_w10_i0.8}]{\includegraphics[width=3in]{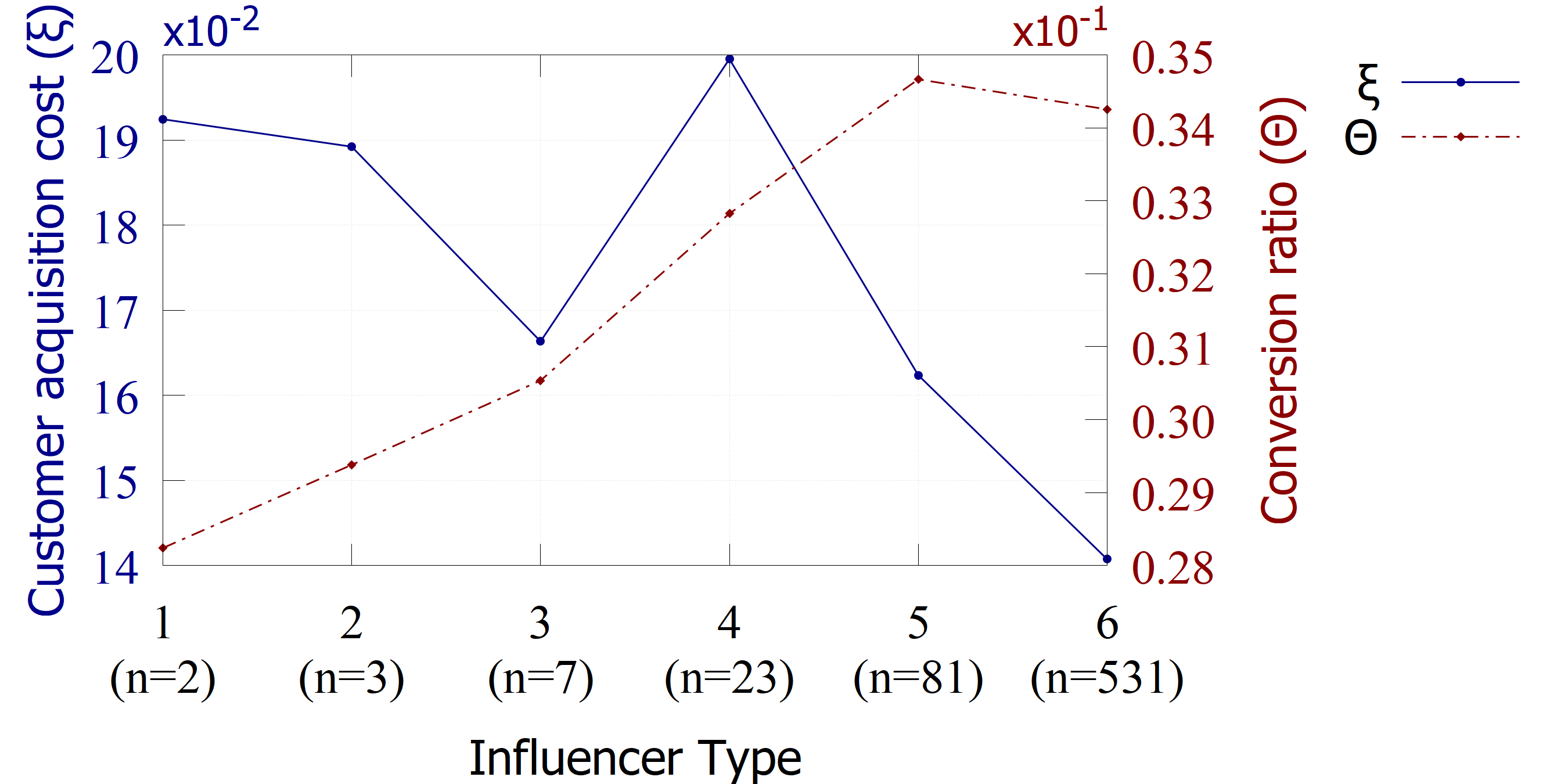}}\qquad
\caption{Simulation results for a luxury product with respect to different types of influencers hired according to the budget $\rho$}
\label{fig:results_luxury_gplus}
\end{figure*}

Similar to the observations made on the Twitter graph in Section \ref{luxury}, we find that for the case of Figure \ref{fig:gplus_w10_i0.5}: (a) Celebrities have a lower customer acquisition cost than nano-influencers. (b) Celebrities have a higher conversion ratio.

In the case of Figure \ref{fig:gplus_w10_i0.8}: (a) Nano-influencers have a lower customer acquisition cost that celebrities. (b) Nano-influencers have a higher conversion ratio than celebrities.

We find that the observations made from experimenting on the real-world Twitter social graph in Section \ref{sec:situational}, for each of the four situations, hold good for Gplus social graph. This provides more certitude on the conclusions drawn on the Twitter social graph. Thus the conclusions drawn are valid in general for social graphs and our model and results can be help marketers make appropriate and informed decisions.

\ifCLASSOPTIONcaptionsoff
  \newpage
\fi

\bibliographystyle{IEEEtran}
\bibliography{references}

\begin{IEEEbiography}
    [{\includegraphics[width=1in,height=1.25in,clip,keepaspectratio]{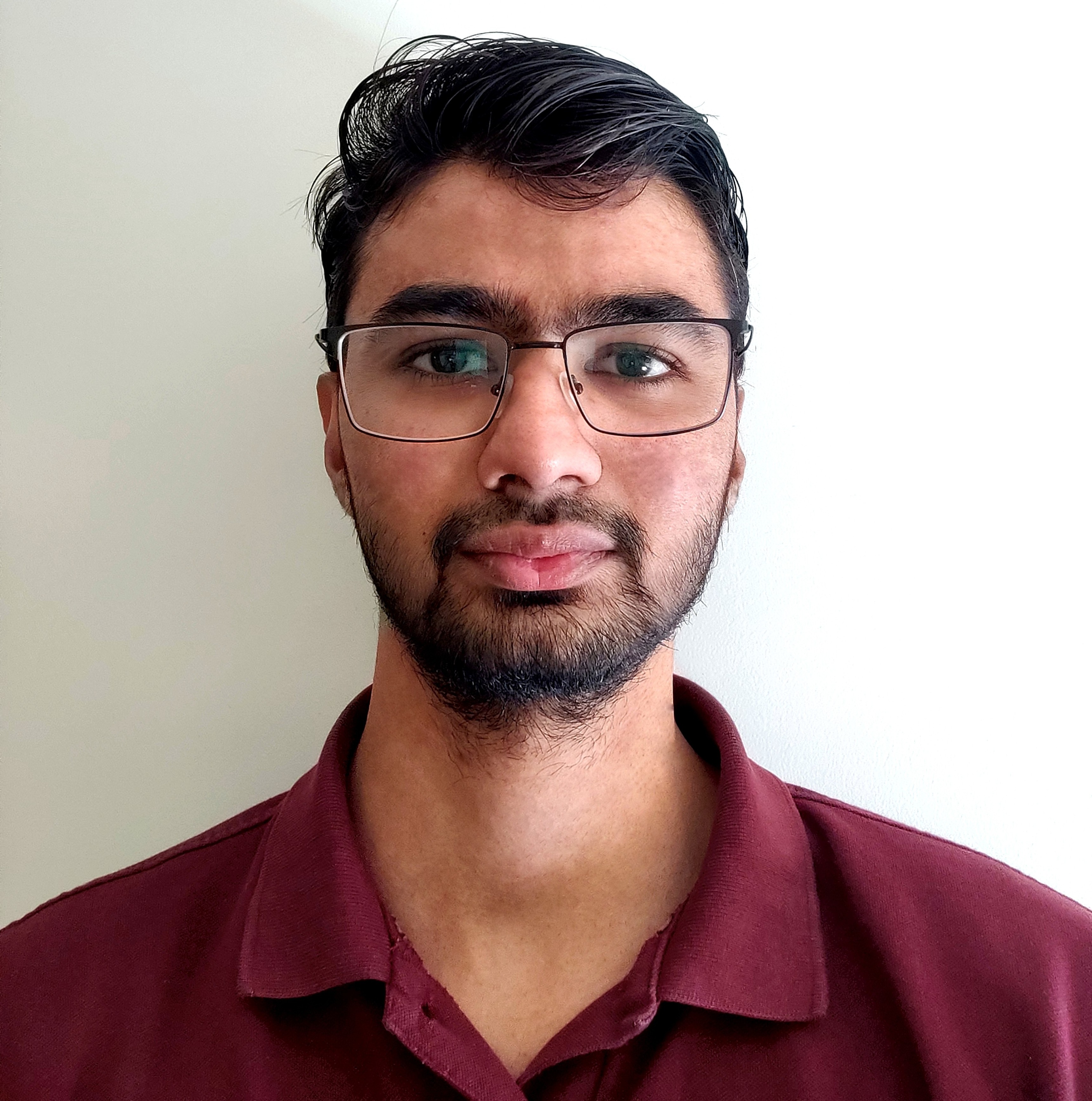}}]{Ronak Doshi}
is currently pursuing the dual B.Tech.
and M.Tech. degrees in electronics and communication,
with specialization in artificial intelligence
and machine learning, with the International Institute
of Information Technology at Bangalore, Bengaluru,
India.

In the past, he has worked with the Neuroinformatics
Laboratory, University of West Bohemia, Pilsen,
Czechia, as a part of the Google Summer of Code
Program, building a web-based toolkit to design
complicated deep learning workflows for EEG signal
processing and classification. He is currently working as a Software Engineering
Intern with Qualcomm, Bengaluru. His research interest lies in applying
the concepts of computer science and programming to real-world applications,
mainly in the field of health care and finance that can benefit society.
\end{IEEEbiography}

\begin{IEEEbiography}
[{\includegraphics[width=1in,height=1.25in,clip,keepaspectratio]{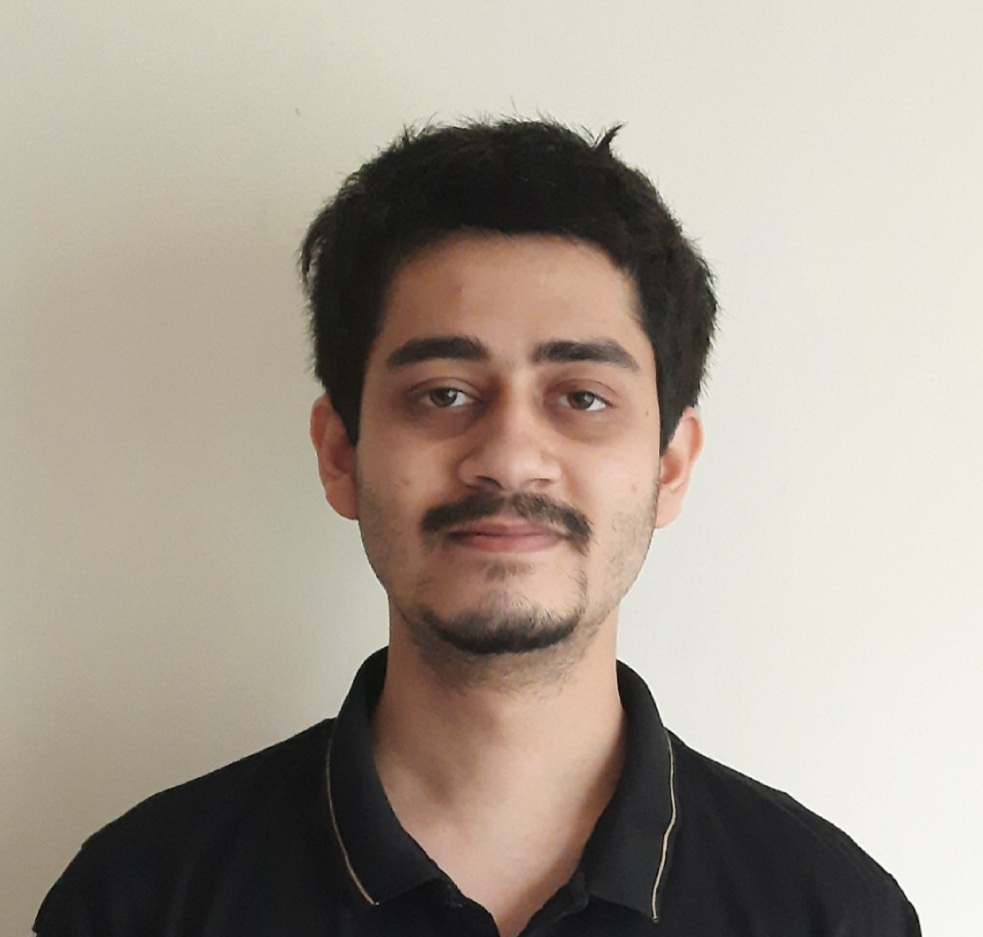}}]{Ajay Ramesh} (Student Member, IEEE) is currently
pursuing the dual B.Tech./M.Tech. degree in
electronics and communication engineering, with a
specialization in artificial intelligence and machine
learning, with the International Institute of Information
Technology at Bangalore, Bengaluru, India.

He is currently an Engineering Development Intern
with MathWorks, Bengaluru. His research experiences
include internships at the Institute of Science
at Bangalore (IISc), Bengaluru, in 2019, and
The University of Alabama, Tuscaloosa, AL, USA,
in 2020. His research interests include artificial intelligence for health care
and assistive technology. He is especially interested in developing machine
learning and computer vision techniques for disease diagnosis and surgical
analysis. He is also interested in healthcare robotics for rehabilitation and
assistance.
\end{IEEEbiography}

\begin{IEEEbiography}
[{\includegraphics[width=1in,height=1.25in,clip,keepaspectratio]{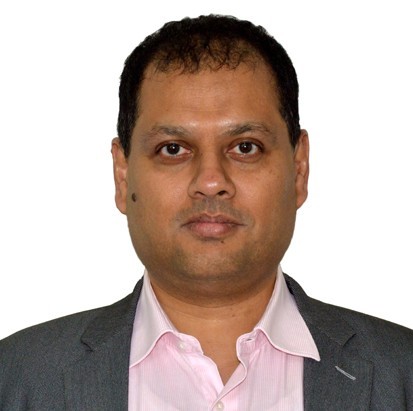}}]{Shrisha Rao}
(Senior Member, IEEE) received the
M.S. degree in logic and computation from Carnegie
Mellon University, Pittsburgh, PA, USA, in 2002,
and the Ph.D. degree in computer science from
The University of Iowa, Iowa City, IA, USA, in
2005.

He is currently a Professor with the International
Institute of Information Technology at Bangalore,
Bengaluru, India. His primary research interests
include artificial intelligence and other applications
of distributed computing, including in bioinformatics
and computational biology, algorithms, and approaches for resource management
in complex systems such as used in cloud computing. He also has
interests in energy efficiency, sustainable computing (“Green IT”), renewable
energy and microgrids, applied mathematics, and intelligent transportation
systems.

Dr. Rao is a Contributing Member of the LITD 14 “Software and System
Engineering” Sectional Committee (a national mirror committee of the ISO
sub-committees IEC/JTC 1/SC 7 Software and System Engineering, and JTC
1/SC 38 Cloud Computing) of the Bureau of Indian Standards (BIS). He is
also a member of the Intelligent Transport Systems Sectional Committee, TED
28, of the BIS.
\end{IEEEbiography}

\end{document}